%% file: GnnDist.tex
\documentclass[acmtog,screen,authorversion]{acmart}
\acmSubmissionID{210}

\usepackage{booktabs}
\usepackage{epsfig}
\usepackage{graphicx}
\usepackage{amsmath}
\usepackage{amssymb}
\usepackage{overpic}
\usepackage{enumitem}
\usepackage{layouts}
\usepackage[capitalise]{cleveref}
\usepackage{subfig, graphicx}
\usepackage{multirow}
\usepackage{multicol}
\usepackage{tikz}
\usepackage{xspace}
\usepackage{wrapfig}
\usepackage[capitalise]{cleveref} 
\usepackage{xcolor}

\usetikzlibrary{intersections}
\graphicspath{{./images/}}

\definecolor{Highlight}{HTML}{39b54a} 
\newcommand{\hl}[1]{\textcolor{Highlight}{#1}}

\newcommand{\tablestyle}[2]{\setlength{\tabcolsep}{#1}
                            \renewcommand{\arraystretch}{#2}
                            \centering
                            \footnotesize}

\def \etal {{\emph{et al}.\thinspace}}

\acmJournal{TOG}
\acmYear{2024}
\acmVolume{42}
\acmNumber{6}
\acmArticle{233}
\acmMonth{12}
\acmPrice{15.00}
\acmDOI{10.1145/3618317}
\setcopyright{acmlicensed}

\received{May 2024}
\received[accepted]{July 2024}
\received[final version]{September 2024}

\title{Learning the Geodesic Embedding with Graph Neural Networks}

\author{Bo Pang}
\affiliation{
  \institution{Peking University}
  \country{China}
}
\email{bo98@stu.pku.edu.cn}
\authornote{Joint first authors.}

\author{Zhongtian Zheng}
\affiliation{
  \institution{Peking University}
  \country{China}
}
\email{zhengzhongtian@pku.edu.cn}
\authornotemark[1]

\author{Guoping Wang}
\affiliation{
  \institution{Peking University}
  \country{China}
}
\email{wgp@pku.edu.cn}

\author{Peng-Shuai Wang}
\affiliation{
  \department{Wangxuan Institute of Computer Technology}
  \institution{Peking University}
  \country{China}
}
\email{wangps@hotmail.com}
\authornote{Corresponding author.}

\begin{abstract}
  We present \textsc{GeGnn}, a learning-based method for computing the approximate geodesic distance between two arbitrary points on discrete polyhedra surfaces with constant time complexity after fast precomputation.
  Previous relevant methods either focus on computing the geodesic distance between a single source and all destinations, which has linear complexity at least or require a long precomputation time.
  Our key idea is to train a graph neural network to embed an input mesh into a high-dimensional embedding space and compute the geodesic distance between a pair of points using the corresponding embedding vectors and a lightweight decoding function.
  To facilitate the learning of the embedding, we propose novel graph convolution and graph pooling modules that incorporate local geodesic information and are verified to be much more effective than previous designs.
  After training, our method requires only one forward pass of the network per mesh as precomputation. 
  Then, we can compute the geodesic distance between a pair of points using our decoding function, which requires only several matrix multiplications and can be massively parallelized on GPUs.
  We verify the efficiency and effectiveness of our method on ShapeNet and demonstrate that our method is faster than existing methods by orders of magnitude while achieving comparable or better accuracy.
  Additionally, our method exhibits robustness on noisy and incomplete meshes and strong generalization ability on out-of-distribution meshes. 
  The code and pretrained model can be found on \url{https://github.com/IntelligentGeometry/GeGnn}.
\end{abstract}

 \begin{teaserfigure}
   \label{tease_fig}
   \centering
   \includegraphics[width=0.96\linewidth]{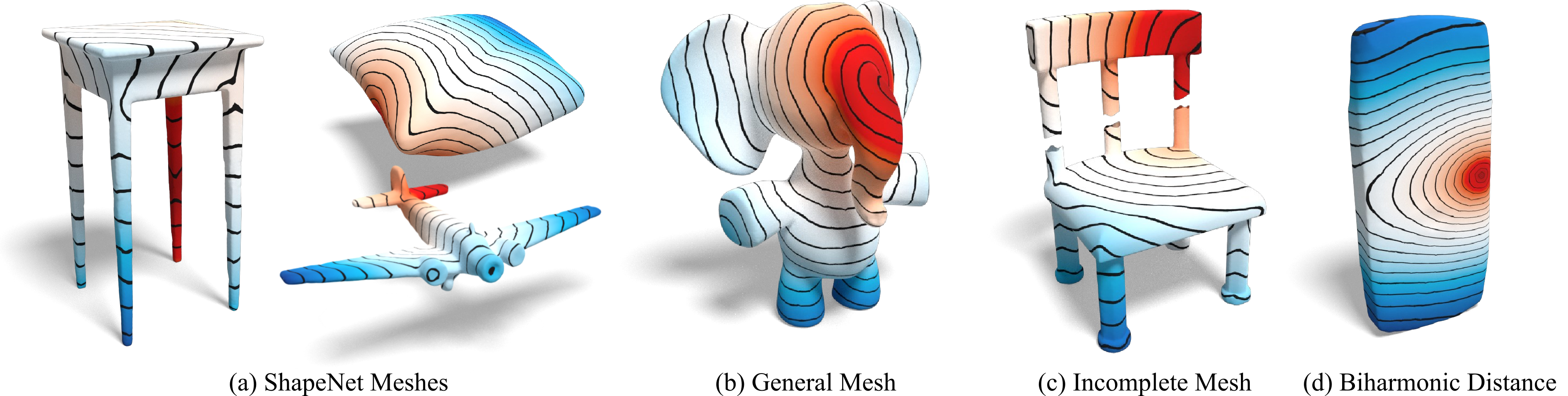}
   \caption{
     Appealing results produced by our method. 
    The distance fields are color-coded using a gradient ranging from red to blue.
    (a) We train a graph neural network on ShapeNet to predict the geodesic distances among arbitrary pairs of points on meshes with constant time complexity after a single forward pass as precomputation; the shown results are from the testing set and unseen categories of ShapeNet with an average relative error of less than 1.4\% and 2.6\%, respectively, which are visually indistinguishable from the ground truth.
    (b) The network exhibits strong generalization capabilities and can be applied to general meshes out of ShapeNet.
    (c) The network demonstrates strong robustness on meshes with corrupted topology, which traditional geodesic algorithms cannot handle.
    (d) Our method is general and can also be used to predict biharmonic distances.
   }
   \label{fig:teaser}
 \end{teaserfigure}



\begin{document}

\maketitle
\input{src/introduction}
\input{src/relatedwork}
\input{src/method}
\input{src/results}
\input{src/conclusion}

\begin{acks}
This work is supported by the \grantsponsor{nkpc}{National Key R\&D Program of China}{} (No.~\grantnum{nkpc}{2022YFB3303400} and No.~\grantnum{nkpc}{2021YFF0500901}),
and the \grantsponsor{nsfc}{Southern Marine Science and Engineering Guangdong Laboratory (Zhuhai)}{} (No.\grantnum{nsfc}{SML2021SP101}).
We also thank the anonymous reviewers for their valuable feedback and Mr. Tianzuo Qin from the University of Hong Kong for his discussion on graph neural networks.
\end{acks}

\bibliographystyle{ACM-Reference-Format}
\bibliography{src/ref/reference}


\end{document}

%% file: src/introduction.tex
\section{Introduction} \label{sec:intro}

The computation of the geodesic distance between two arbitrary points on polyhedral surfaces, also referred to as the geodesic distance query (GDQ), is a fundamental problem in computational geometry and graphics and has a broad range of applications, including texture mapping~\cite{Zigelman2002}, symmetry detection~\cite{Xu2009}, mesh deformation~\cite{Bendels2003}, and surface correspondence~\cite{Raviv2010}.
Although plenty of methods~\cite{Mitchell1987,Ying2013,Adikusuma2020,Crane2013a} have been proposed for computing single-source-all-destinations geodesic distances, and some of them can even run empirically in linear time~\cite{Crane2013a,Ying2013,Tao2019}, it is still too expensive to leverage these methods for GDQs.
Consequently, a few dedicated methods~\cite{Xin2012,Panozzo2013,Xia2021,Gotsman2022} have been proposed to ensure that the computation complexity of GDQ is constant to meet the huge requirements of frequent GDQs in interactive applications.

However, these methods either have low accuracy or require long precomputation time, which severely limits their applicability to large-scale meshes.
Specifically, in the precomputation stage, these methods typically need to compute the exact geodesic distances between a large number of vertex pairs~\cite{Gotsman2022,Xin2012,Panozzo2013,Xia2021}, incurring at least quadratic space and computation complexity;
then some methods employ nonlinear optimization~\cite{Panozzo2013,Rustamov2009} like Metric Multidimensional Scaling (MDS)~\cite{Carroll1998} or cascaded optimization~\cite{Xia2021} to embed the input mesh into a high-dimensional space and approximate the geodesic distance with Euclidean distance in the high-dimensional space, where the optimization process is also time-consuming and costs several minutes, up to hours, even for meshes with tens of thousands of vertices.
Additionally, these methods are highly dependent on the quality of input meshes, severely limiting their ability to deal with noisy or incomplete meshes.

In this paper, we propose a learning-based method for const time GDQ on arbitrary meshes.
Our key idea is to train a graph neural network (GNN) to embed an input mesh into a high-dimensional feature space and compute the geodesic distance between any pair of vertices with the corresponding feature distance defined by a learnable mapping function.
The high-dimensional feature space is referred to as a geodesic embedding of a mesh~\cite{Xia2021,Panozzo2013}.
Therefore, our method can be regarded as learning the geodesic embedding with a GNN, instead of relying on costly optimization procedures~\cite{Panozzo2013,Rustamov2009,Xia2021}; thus, we name our method as \textsc{GeGnn}.
After training, our \textsc{GeGnn} can predict the geodesic embedding by just one forward pass of the network on GPUs, which is significantly more efficient than previous optimization-based methods~\cite{Xia2021,Panozzo2013,Rustamov2009}.
Our \textsc{GeGnn} also learns shape priors for geodesic distances during the training process, which makes it robust to the quality of input meshes.
As a result, our \textsc{GeGnn} can even be applied to corrupted or incomplete meshes, with which previous methods often fail to produce reasonable results.

The key challenges of our \textsc{GeGnn} are how to design the graph convolution and pooling modules, as well as the mapping function for geodesic distance prediction.
Although plenty of graph convolutions have been proposed~\cite{Wu2020} and widely used for learning on meshes~\cite{Hanocka2019,Fey2018}, they are mainly designed for mesh understanding, thus demonstrating inferior performance for our purpose.
Our key observation is that the local geometric structures of a mesh are crucial for geodesic embedding.
To this end, we propose a novel graph convolution by incorporating local distance features on edges and an adaptive graph pooling module by considering the normal directions of vertices.
Our graph convolution follows the message-passing paradigm~\cite{Gilmer2017,Simonovsky2017}, which updates the vertex feature by aggregating neighboring vertex features. 
Inspired by the Dijkstra-like methods for computing geodesic distances~\cite{Tsitsiklis1995,Sethian1999} that propagates extremal distance on the wavefront, we propose aggregating local features with the \emph{max} operator instead of \emph{sum} or \emph{mean}, which we find to be much more effective for geodesic embedding. \looseness=-1

After embedding an input mesh into a high-dimensional feature space with the proposed GNN, one may naively follow previous works~\cite{Panozzo2013,Rustamov2009} to use the Euclidean distance in the embedding space between two vertices to approximate the geodesic distance.
However, we observe that the Euclidean distance cannot well approximate the geodesic distance, regardless of the dimension of the embedding space.
To tackle this issue, we propose to use a lightweight multilayer perceptron (MLP) as a learnable distance function to map the embedding features of a pair of vertices to the geodesic distance, which turns out to work much better than the Euclidean distance.

We train our \textsc{GeGnn} on ShapeNet and verify its effectiveness and efficiency over other state-of-the-art methods for GDQs.
Our \textsc{GeGnn} has constant time complexity after a single forward pass of the network as precomputation and linear space complexity, which is much more efficient than previous methods~\cite{Xia2021,Panozzo2013,Rustamov2009}, with a speedup of orders of magnitude for precomputation and comparable approximation errors for geodesic distances.
Our \textsc{GeGnn} also demonstrates superior robustness and strong generalization ability during the inference stage.
Finally, we showcase a series of interesting applications supported by our \textsc{GeGnn}. 

In summary, our main contributions are as follows:
\begin{itemize}[leftmargin=12pt,itemsep=2pt]
  \item[-] We propose \textsc{GeGnn}, a learning-based method for GDQ using a graph neural network with constant time complexity after a single forward pass as precomputation.
  \item[-] We propose a novel graph convolution module and a graph pooling module, which significantly increase the capability of \textsc{GeGnn} on learning the geodesic embedding.
  \item[-] We propose to use a learnable distance function to map the embedding features of a pair of vertices to the geodesic distance, resulting in a significant improvement in the approximation accuracy compared with the Euclidean distance.
  \item[-] We conduct experiments to demonstrate that the proposed framework is robust and generalizable and is also applicable to predict the biharmonic distance.
\end{itemize}

%% file: src/relatedwork.tex
\section{Related Work} \label{sec:related}

\begin{figure*}[ht]
    \centering
    \includegraphics[width=0.9\linewidth]{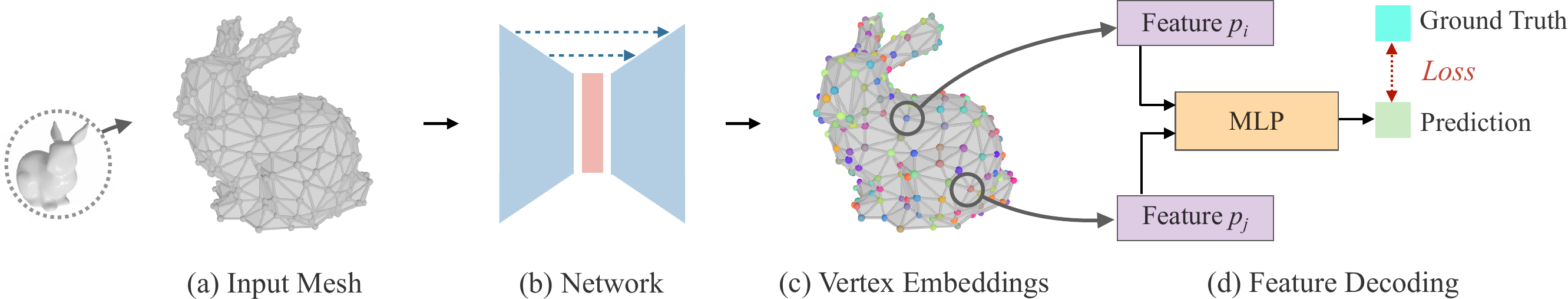}
    \caption{Overview of our \textsc{GeGnn}.
    We construct a graph from an input mesh (a), and feed the graph to a graph neural network (b). The network architecture follows U-Net.
    The network lifts every vertex to a high-dimensional embedding space (c).
    Then we use a lightweight MLP to decode the embeddings of a vertex pair to the corresponding geodesic distance (d).
    In the training stage, we obtain a loss by comparing the predicted geodesic distance with the precomputed ground truth geodesic distance, which is used to backpropagate through the network and update its parameters. In the testing stage, we can directly use the trained network to predict the vertex features and then compute geodesic distance between any two vertices on the mesh with the MLP.    }
    \label{fig:overview}
\end{figure*}

\paragraph{Single-Source Geodesic Distances}
Plenty of algorithms have been developed for the problem of single source geodesic distances computing, we refer the readers to~\cite{Crane2020} for a comprehensive survey.
Representative methods include wavefront-propagation-based methods with a priority queue following the classic Dijkstra algorithm~\cite{Mitchell1987,Chen1990,Surazhsky2005,Xin2009,Ying2014,Xu2015b,Qin2016}, PDE-based methods by solving the Eikonal equation~\cite{Sethian1999,Kimmel1998,Memoli2001,Memoli2005,Crane2013,Crane2017}, and Geodesic-graph-based methods~\cite{Ying2013,Adikusuma2020,Sharp2020}.
There are also parallel algorithms that leverage GPUs for acceleration~\cite{weber2008parallel}.
However, the computational cost of these methods is at least linear to the number of vertices,  while our method has constant complexity during inference.

\paragraph{Geodesic Distance Queries}
The goal of our paper is to approximate the geodesic distance between two arbitrary vertices on a mesh in const time after fast precomputation.
The research on this problem is relatively preliminary, and there are only a few related works~\cite{Xin2012,Panozzo2013,shamai2018efficient,Gotsman2022,zhang2023neurogf}.
A naive solution is to compute the geodesic distance between all pairs of vertices in advance and store them in a lookup table.
However, this method is not scalable to large meshes due to its quadratic time and space complexity.
Xin \etal~\shortcite{Xin2012} propose to construct the geodesic Delaunay triangulation for a fixed number of sample vertices, compute pairwise geodesic distances, record the distances of each vertex to the three vertices of the corresponding geodesic triangle, and approximate the geodesic distance between two arbitrary vertices using triangle unfolding.
Its time and space complexity is quadratic to the number of samples, which is still not scalable to large meshes.
Panozzo \etal~\shortcite{Panozzo2013} propose to embed a mesh into a high-dimensional space to approximate the geodesic distance between a pair of points with the Euclidean distance in the embedding space, which is realized by optimizing with Metric MDS~\cite{Carroll1998,Rustamov2009}.
However, the metric MDS in \cite{Panozzo2013} involves a complex nonlinear optimization and thus is computationally expensive.
Shamai \etal~\shortcite{shamai2018efficient} solve pairwise geodesics employing a fast classic MDS, which extrapolates distance information obtained from a subset of points to the remaining points.
Xia \etal~\shortcite{Xia2021} propose to compute the geodesic embedding by cascaded optimization on a proximity graph containing a subset of vertices~\cite{Ying2013}.
Likewise, the computational expense of this optimization process makes it impractical for large meshes.
Recently, Gotsman \etal~\shortcite{Gotsman2022} propose to compress the pairwise geodesic distances by a heuristic divide and conquer algorithm, but its worst case time complexity for GDQ is not constant.
In contrast, our \textsc{GeGnn} needs no expensive precomputation and has remarkable efficiency for GDQ, thus is scalable to large meshes.
Concurrently, Zhang \etal~\shortcite{zhang2023neurogf} also propose to use neural networks for GDQs; however, this method concentrates on learning on a single mesh, while our method is trained on a large dataset to achieve the generalization capability across different meshes.

\paragraph{Other Geometric Distances}
Apart from the geodesic distance, there are also several other distances on meshes.
The diffusion distance~\cite{Coifman2005} is defined by a diffusion process on the mesh and is widely used for global shape analysis.
The commute-time distance~\cite{Fouss2007,Yen2007} is intuitively described by the expected time for a random walk to travel between two vertices.
The biharmonic distance~\cite{Lipman2010} can be calculated by the solution of the biharmonic equation.
The earth mover's distance~\cite{Solomon2014a} is defined by the minimum cost of moving the mass of one point cloud to another.
Apart from geodesic distances, our \textsc{GeGnn} can also be easily extended to approximate these distances by replacing the geodesic distance with the corresponding distance in the loss function. We showcase the results of approximating the biharmonic distance in our experiments.

\paragraph{Graph Neural Networks}
Graph neural networks (GNN) have been widely used in many applications~\cite{Wu2020}.
As the core of a GNN, a graph convolution module can be formulated under the message passing paradigm~\cite{Gilmer2017,Simonovsky2017}.
Representative GNNs include the graph convolutional network~\cite{Kipf2017}, the graph attention network~\cite{Velickovic2017}, and GraphSAGE~\cite{Hamilton2017}.
Many point-based neural networks for point cloud understanding~\cite{Qi2017,Li2018,Thomas2019,Xu2018,Wang2019c} can be regarded as special cases of graph neural networks on k-nearest-neighbor graphs constructed from unstructured point clouds.
Graph neural networks can also be applied to triangle meshes for mesh understanding~\cite{Hanocka2019,Hu2022,Yi2017b}, simulation~\cite{Pfaff2020}, processing~\cite{Liu2020c}, and generation~\cite{Hanocka2020}.
However, these methods are not specifically designed for GDQs and demonstrate inferior performance compared to our dedicated graph convolution in terms of approximation accuracy. 

%% file: src/method.tex
\section{Method} \label{sec:method}

\paragraph{Overview}
The overall pipeline of our \textsc{GeGnn} is shown in \cref{fig:overview}.
Our method can be roughly separated into two parts: precomputation, and geodesic distance query (GDQ) . 
The precompuation step (\cref{fig:overview} (b)) involves evaluating a graph neural network to obtain vertex-wise features of fixed dimension. This step is performed only once for each mesh, regardless of the number of subsequent queries.
The geodesic distance query (\cref{fig:overview} (d)) involves forwarding the powered difference between two features to a fixed size MLP. This process requires a fixed number of max/add/multiplication operations for each query, resulting in a time complexity of $O(1)$ for GDQ.
Specifically, given an input mesh $\mathcal{M}=\{\mathcal{V},\mathcal{F}\}$, where $\mathcal{V}$ and $\mathcal{F}$ denote the set of vertices and faces of the mesh, the connectivity of the mesh defined by $\mathcal{F}$ naturally forms a graph $\mathcal{G}$.
We first train a graph neural network that takes $\mathcal{G}$  and $\mathcal{V}$ as input and maps each vertex $v_i$ in $\mathcal{V}$ to a high dimensional embedding vector $p_i \in \mathbb{R}^c$, where $c$ is set to 256 by default.
We construct a U-Net~\cite{Ronneberger2015} on the graph to effectively extract features for each vertex.
Then we use a lightweight MLP to take the features of two arbitrary vertices as input and output the corresponding geodesic distance between them.
The key building blocks of our network include a novel graph convolution module and a graph pooling module, which are elaborated in \cref{subsec:gnn}.
The feature decoding scheme for geodesic distance prediction is elaborated in \cref{subsec:decode}.
Finally, the network details and the loss function are introduced in \cref{subsec:detail}.  \looseness=-1

\subsection{Graph Neural Network} \label{subsec:gnn}

In this section, we introduce our graph convolution and pooling modules tailored for learning the geodesic embedding, which serve as the key building blocks of our network.

\subsubsection{Graph Convolution}  \label{subsec:conv}
The graph convolution module is used to aggregate and update features on a graph.
The graph $\mathcal{G}$ constructed from an input mesh $\mathcal{M}=\{\mathcal{V},\mathcal{F}\}$ contains the neighborhood relationships among vertices.
For the $i^{th}$ vertex $v_i$ in $\mathcal{V}$, we denote the set of its neighboring vertices as $\mathcal{N}(i)$ and its feature as $F_i$.
Generally, denote the output of the graph convolution module as $F_i^{\prime}$, the module under the message passing paradigm~\cite{Gilmer2017,Simonovsky2017,Fey2019}  can be defined as follows:
\begin{equation}
  F_i^{\prime} = \gamma \left(F_i, \; \mathcal{A}_{j \in \mathcal{N}(i)} \phi \left(F_i, F_j, v_i, v_j\right)  \right), 
\end{equation}
where $\gamma$ and $\phi$ are differentiable functions for updating features, 
and $\mathcal{A}$ is a differentiable and permutation invariant function for aggregating neighboring features, e.g., \emph{sum}, \emph{mean}, or \emph{max}.
The key difference among different graph convolution operators lies in the design of $\gamma$, $\phi$, and $\mathcal{A}$.

Although plenty of graph convolutions have been proposed, they are mainly used for graph or mesh understanding~\cite{Fey2018,Velickovic2017,Kipf2017} and are insensitive to the local distances between vertices, which are crucial for geodesic embedding.
Our design philosophy is to explicitly incorporate the distances between neighboring vertices into the graph convolution, with a focus on maximizing simplicity while maintaining expressiveness.
Therefore, we define the functions for updating features $\phi$ and $\gamma$ as follows:
\begin{align} 
  \gamma(F_i, \bar{F}_i) &=  W_0 \times F_i + \bar{F}_i,  \\
  \phi \left(F_i, F_j, v_i, v_j\right) &=  W_1   \times [F_j \mathbin\Vert v_{ij} \mathbin\Vert l_{ij} ],
\end{align}
where $\bar{F}_i = \mathcal{A}_{j \in \mathcal{N}(i)} \phi \left(F_i, F_j, v_i, v_j\right)$, $W_0$ and $W_1$ are trainable weights, $\mathbin\Vert$ is a concatenation operator, $v_{ij}$ is a shorthand for $v_i - v_j$, and $l_{ij}$ is the length of $v_{ij}$.
For the aggregation operator, we choose \emph{max}, instead of \emph{sum} or \emph{mean}, which is motivated by Dijkstra-like methods for computing geodesic distances~\cite{Tsitsiklis1995,Sethian1999} that propagate the shortest distance on the wavefront.
This choice is also consistent with the observation in~\cite{Xu2018a} that the \emph{max} is advantageous in identifying representative elements.
In summary, our graph convolution is defined as follows:
\begin{equation}
  F_i^{\prime} = W_0 \times F_i +   \max\nolimits_{j \in \mathcal{N}(i)} \left( W_1 \times [F_j \mathbin\Vert v_{ij} \mathbin\Vert l_{ij}] \right),
  \label{eq:conv}
\end{equation}
Since our graph convolution is designed for geodesic embedding, we name it as \emph{GeoConv}.
According to \cref{eq:conv}, our \emph{GeoConv} does not use any global information such as absolute positions, and thus is
translation and permutation invariant by design, which are desired properties for geodesic computing.

Compared with previous graph convolutions in the field of 3D deep learning~\cite{Wang2019c,Simonovsky2017} that define $\phi$ or $\gamma$ as MLPs, our \emph{GeoConv} is much simpler, while being more effective for geodesic embedding, as verified in our experiments and ablation studies.
Our \emph{GeoConv} is reminiscent of GraphSAGE~\cite{Hamilton2017} for graph node classification on citation and Reddit post data; however, GraphSAGE does not consider local distances between vertices and uses \emph{mean} for aggregation, resulting in an inferior performance for geodesic embedding.

\begin{figure}[t]
  \centering
  \includegraphics[width=\linewidth]{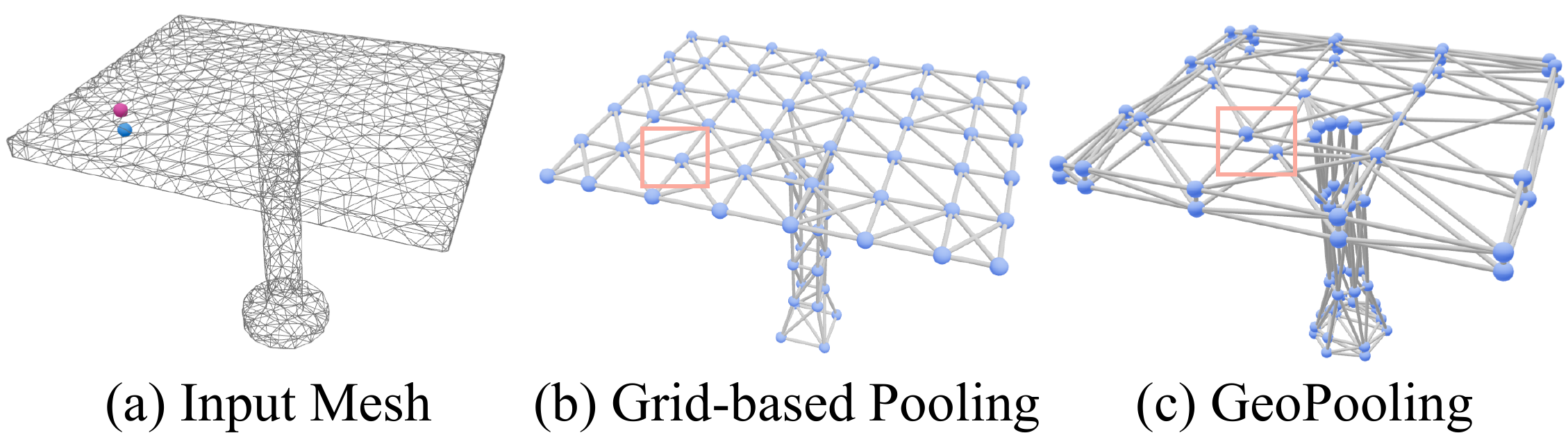}
  \caption{ Comparison of graph pooling strategies.
    (a) The red and blue vertices are located on opposite sides of the tabletop. Although they are distant in terms of geodesic distance, they are close in Euclidean space.
    (b) The grid-based pooling merges the red and blue vertices into a single vertex, which is inadequate for geodesic embedding.
    (c) Our \emph{GeoPool} is capable of preserving the geodesic distance and effectively distinguishing between the red and blue vertices.}
\label{fig:pooling}
\end{figure}

\subsubsection{Graph Pooling} \label{subsec:pool}
Graph pooling is used to progressively downsample the graph to facilitate the construction of U-Net.
Representative graph pooling designs in previous works are either based on mesh simplification with a priority queue~\cite{Hanocka2019} or leverage farthest point sampling~\cite{Qi2017}, incurring huge computational cost and diffculties concerning parallelism.
Therefore, we resort to grid-based graph pooling for efficiency~\cite{Hu2019,Thomas2019,Simonovsky2017}.
Grid-based pooling employs regular voxel grids to cluster vertices within the same voxel into a single vertex, resulting in a straightforward and efficient way to downsample the graph.
Nevertheless, grid-based pooling only relies on the Euclidean distance of vertices and ignores the topology of the underlying graph or mesh.
An example is shown in \cref{fig:pooling}-(a): the red and blue vertices are on the opposite sides of the tabletop; they are far away in terms of geodesic distance but close in Euclidean space.
Grid-based pooling may merge these two vertices into a single vertex as shown in \cref{fig:pooling}-(b), which is undesirable for geodesic embedding. 

To address this issue, we propose a novel graph pooling, called \emph{GeoPool}, which is aware of both Euclidean and geodesic distances.
Specifically, we construct regular grids in 6D space consisting of vertex coordinates and the corresponding normals. 
We specify two scale factors, $\sigma_c$ for the coordinates and $\sigma_n$ for the normals, to control the size of the grids in different dimensions. 
Then we merge vertices within the same grid in 6D space and compute the averages of coordinates, normals, and associated features as the output.
The use of normals can effectively prevent vertices on the opposite sides of a thin plane from being merged when downsampling the graph.
As shown in \cref{fig:pooling}-(c), the red and blue vertices are in different grids in 6D space, and thus are not merged into a single vertex with \emph{GeoPool}.
Note that \emph{GeoPool} reduces to vanilla grid-based pooling when $\sigma_n$ is set to infinity.

We also construct a graph unpooling module, named \emph{GeoUnpool}, for upsampling the graph in the decoder of U-Net.
Specifically, we keep track of the mapping relation between vertices before and after the pooling operation, and \emph{GeoUnpool} is performed by reversing the \emph{GeoPool} with the cached mapping relation.

\subsubsection{Network Architecture} \label{subsec:arch}
\begin{figure}[t]
  \centering
  \includegraphics[width=\linewidth]{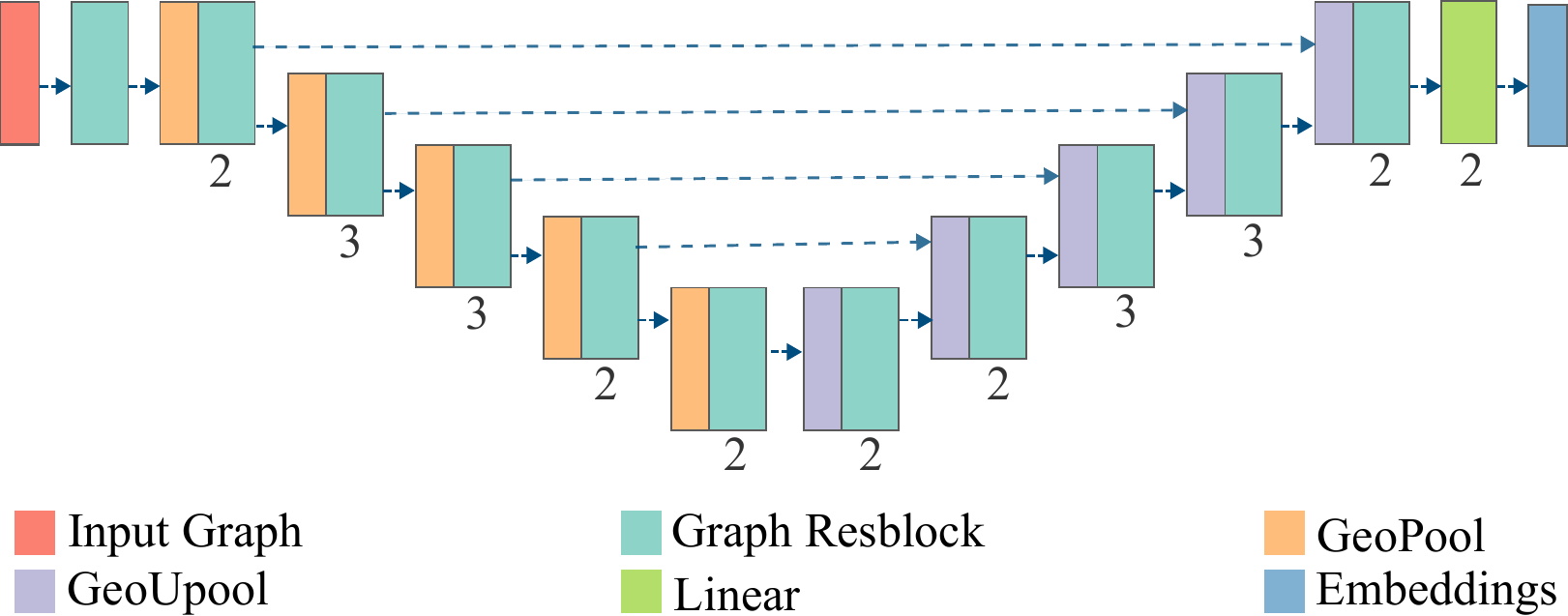}
  \caption{U-Net constructed with our basic network modules. The network takes a graph as input and outputs high-dimensional geodesic embeddings for vertices.
  The feature channels are all set to 256, and the numbers of each block are  shown underneath. }
\label{fig:network}
\end{figure}

We construct a U-Net~\cite{Ronneberger2015} with our \emph{GeoConv}, \emph{GeoPool}, and \emph{GeoUnpool} modules, as shown in \cref{fig:network}.
The network takes the graph constructed from an input mesh as input.
The initial vertex signals have 6 channels, including the normalized vertex coordinates and normals.
The vertex features are updated by \emph{GeoConv}; and each \emph{GeoConv} is followed by a ReLU activation and a group normalization~\cite{Wu2018d}.
The \emph{ResBlock} in \cref{fig:network} is built by stacking 2 \emph{GeoConv} modules with a skip connection~\cite{He2016}.
The output channels of each \emph{GeoConv} are all set to 256. 
The graph is progressively downsampled and upsampled by \emph{GeoPool} and \emph{GeoUnpool}.
Since the input mesh is first normalized into $[-1, 1]$,
$\sigma_c$ and $\sigma_n$ are initially set to $1/16$ and $3/16$, and increase by a factor of 2 after each pooling operation.
Overall, the U-Net extracts a feature for every vertex, which is then mapped to 256 channels as the geodesic embedding using a MLP consisting of two fully-connected layers with 256 channels.

The geodesic distance between two arbitrary vertices is determined by the shortest path between them, which is a global property of the mesh.
The U-Net built upon our graph modules has global receptive fields, while being aware of local geometric structures, which is advantageous for geodesic embedding.
The multi-resolution structure of U-Net also increases the robustness when dealing with incomplete meshes, since the missing edges or isolated vertices in incomplete meshes can be merged and get connected in coarser resolutions.
We verify the robustness of our network in the experiments. \looseness=-1

\subsection{Geodesic Decoder}\label{subsec:decode}

The graph network outputs a geodesic embedding $p_i \in \mathbb{R}^{256}$ for each vertex $v_i$.
To approximate the geodesic distance between vertex $v_i$ and $v_j$, we need to define a distance function $d(p_i, p_j)$ in the embedding space as the decoder.
An intuitive strategy is to directly use the Euclidean distance between $p_i$ and $p_j$, following~\cite{Panozzo2013,Rustamov2009}: $d(p_i, p_j) = || p_i - p_j||_2$.
However, such an Euclidean embedding is unattainable in most scenarios, including curved surfaces with non-zero Gaussian curvature~\cite{pressley2010gauss} and finite metric spaces with Gramian matrices that lack positive semidefiniteness~\cite{maehara2013euclidean}. 

\begin{wrapfigure}[8]{r}{0.28\linewidth}
  \centering
  \vspace{-6pt}
  \hspace{-10pt}
  \includegraphics[width=\linewidth]{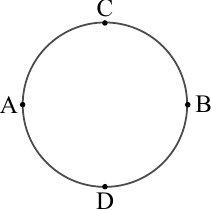}
\end{wrapfigure}
Here is a simple example to illustrate the issue. 
Denote the geodesic distance between point $X$ and point $Y$ as $d_{XY}$.
For a unit circle depicted on the right, we can easily compute that $d_{AC} = d_{BC} = \pi/2$ and $d_{AB}=\pi$.
If we embed this circle into a high-dimensional Euclidean space $\mathcal{X}$, we will have $C$ at the midpoint of the segment $AB$ in $\mathcal{X}$.
Similarly, when considering $A$, $B$, and $D$, we know that $D$ is also the midpoint of the segment $AB$.
Therefore, $C$ and $D$ would be at the same point in $\mathcal{X}$; however, $d_{CD}$ is equal to $\pi$ on the circle.
This example shows that the Euclidean distance cannot well approximate the geodesic distance even on a circle, which is irrelevant to the dimension of the embedding space.

Previous efforts have also attempted manual design of embedding spaces~\cite{shamai2018efficient}, which involves mapping data onto a sphere.
Instead of manually designing the decoding function $d(p_i, p_j)$, which turns out to be tedious and hard according to our initial experiments, we propose to learn it from data.
Specifically, we train a small MLP with 3 fully connected layers and ReLU activation functions in between to learn the decoding function, as shown in \cref{fig:overview}-(d).
The channels of the decoding MLP are set to 256, 256, and 1, respectively.
Define $p_i = \left(p_{i,1}, \; p_{i,2}, \; \dots, \; p_{i,n}\right)$,
where  $p_{i,k}$ is the $k^{th}$ components of $p_i$ and $n$ is the dimension of $p_i$, 
the MLP takes the squared difference $s_{ij}$ between $p_i$ and $p_j$ as the input:
\begin{align} 
  s_{ij} = \left( \left(p_{i,1} - p_{j,1}\right)^2, \;  \left(p_{i,2} - p_{j,2}\right)^2, \; \dots, \; \left(p_{i,n} - p_{j,n}\right)^2 \right),
  \label{eq:decoder}
\end{align}
and the output of the MLP approximates the geodesic distance between $v_i$ and $v_j$: $d(p_i, p_j) = MLP(s_{ij})$.
The square operation in \cref{eq:decoder} is necessary as it preserves the symmetry of $d(p_i, p_j)$  with respect to $p_i$ and $p_j$.

Since the MLP only contains 3 layers, its execution on GPUs is very efficient.
For each input mesh, we forward the U-Net once to obtain the geodesic embedding of each vertex and cache the results; then each batch of GDQs only requirs a single evaluation of the MLP. \looseness=-1

\subsection{Loss Function} \label{subsec:detail}

We adopt the mean relative error (MRE) as the loss function for training, which is also used as the evaluation metric to assess the precision of predicted geodesic distances following~\cite{Surazhsky2005,Xia2021}.

In the data preparation stage, we precompute and store $N$ pairs of exact geodesic distances for each mesh with the method proposed in~\cite{Mitchell1987}.
In each training iteration, we randomly sample a batch of meshes from the training set, then randomly sample $n$ pair of vertices from the precomputed geodesic distances for each mesh.
The MRE for each mesh can be defined as follows:
\begin{equation}
  L= \frac{1}{n} \sum\nolimits_{(i,j) \in \mathcal{I}} \frac{|d(p_i, p_j)-d_{ij}|}{d_{ij}+\epsilon},
  \label{eq:loss}
\end{equation}
where $\mathcal{I}$ is the set of sampled vertex pairs, $d_{ij}$ is the ground-truth geodesic distance between vertex $v_i$ and vertex $v_j$, $d(p_i, p_j)$ is the decoding function described in \cref{subsec:decode}, and $\epsilon$ is a small constant to avoid numerical issues.
In our experiments, we set $N$ to $600k$, $n$ to $100k$, and $\epsilon$ to $0.001$.

\paragraph{Remark}
Although our main goal is to predict the geodesic distance, our method is general and can be trivially applied to learn other forms of distances.
For example, we precompute Biharmonic distances~\cite{Lipman2010} for certain meshes and use them to replace the geodesic distances in \cref{eq:loss} to train the network.
After training, our network can learn to estimate the Biharmonic distances as well.
We verify this in our experiments, and we expect that our method has the potential to learn other types of distances as well, which we leave for future work.  \looseness=-1

%% file: src/results.tex
\section{Results} \label{sec:result}

In this section, we validate the efficiency, accuracy, and robustness of our \textsc{GeGnn} on the task of GDQ and demonstrate its potential applications.
We also analyze and discuss key design choices of \textsc{GeGnn} in the ablation study.
The experiments were conducted using 4 Nvidia 3090 GPUs with 24GB of memory. 

\subsection{Geodesic Distance Queries} \label{subsec:gdq}

\paragraph{Dataset}
We use a subset of ShapeNet~\cite{Chang2015} to train our network.
The subset contains 24,807 meshes from 13 categories of ShapeNet, of which roughly 80\% are used for training and others for testing.
Since the meshes from ShapeNet are non-manifold, prohibiting the computation of ground-truth geodesic distances, we first convert them into watertight manifolds following~\cite{Wang2022}.
Then, we apply isotropic explicit remeshing~\cite{Pravin2004,Hoppe1993} to make the connectivity of meshes regular.
Next, we do mesh simplification~\cite{Garland1997} to collapse 15\% of the edges, which diversifies the range of edge length and potentially increases the robustness of our network.
The resulting meshes have an average of 5,057 vertices.
We normalize the meshes in $[-1, 1]$ and leverage the MMP method~\cite{Mitchell1987} to calculate the exact geodesic distance of $600k$ pairs of points on each mesh, which takes 10-15s on a single CPU core for each mesh.

\begin{figure}[t]
    \centering
    \includegraphics[width=\linewidth]{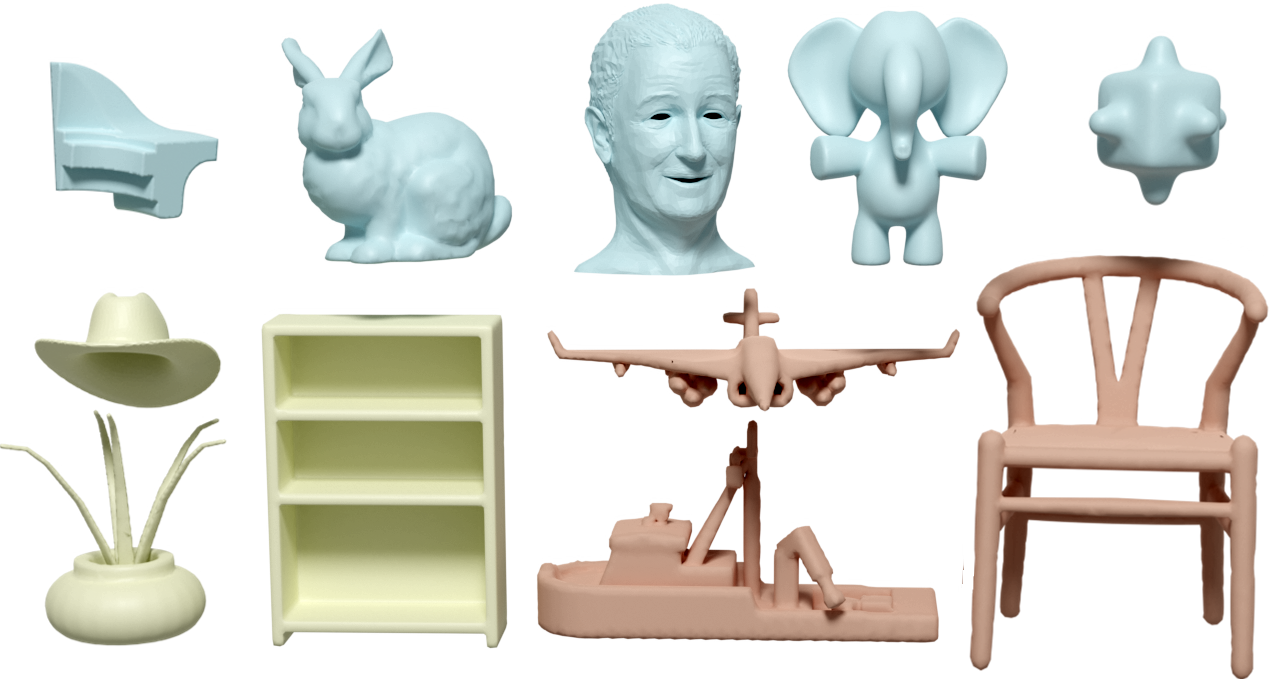}
    \caption{Example meshes for evaluation. The meshes from the ShapeNet-A, ShapeNet-B, and Common groups are shown in red, yellow, and blue, respectively. Notice that our network can handle meshes with boundaries: the head (MRE=2.0\%) has open boundaries near its eyes and neck.}
    \label{fig:meshes}
\end{figure}

\paragraph{Settings}
We employ the AdamW optimizer~\cite{Loshchilov2017} for training, with an initial learning rate of 0.0025, a weight decay of 0.01, and a batch size of 40.
We train the network for 500 epochs and decay the learning rate using a polynomial function with a power of 0.9.
We implemented our method with PyTorch~\cite{Paszke2019}; the training process took 64 hours on 4 Nvidia 3090 GPUs.
We use the mean relative error (MRE) defined in \cref{eq:loss} to compare the accuracy of the predicted geodesic distance.
To compute the MRE, we randomly sample $500$ vertices, calculate the geodesic distances from each vertex and all other vertices within a given mesh, and then compute the average error between the predicted distances and the ground-truth distances.
We use the average time required of 1 million geodesic queries and pre-processing time to evaluate the efficiency.
We choose three groups of meshes for evaluation:
\begin{itemize}[leftmargin=10pt,itemsep=2pt]
    \item[-] ShapeNet-A: 100 meshes from the testing set of ShapeNet, whose categories are included in the training set. The average number of vertices is 5,057.
    \item[-] ShapeNet-B: 50 meshes from the ShapeNet, whose categories are \emph{different} from the 13 categories of the training set.  The average number of vertices is 5,120.
    \item[-] Common: 10 meshes that are not contained in ShapeNet, such as Elephant, Fandisk, and Bunny. The topologies, scales, and geometric features of these meshes are significantly different from those in ShapeNet. The average number of vertices is 7,202.
\end{itemize}
Several examples from these three groups of meshes are shown in \cref{fig:meshes}.
The meshes from ShapeNet-B and Common are used to test the generalization ability of our method when dealing with out-of-distribution meshes.

\paragraph{Comparisons}
We compare our method with several state-of-the-art methods on geodesic distance queries (GDQs).
MMP~\cite{Mitchell1987} is a classic method for computing geodesic distances, with a time complexity of $\mathcal{O}(N^2\log N)$ for each GDQ, where $N$ is the number of mesh vertices.
MMP does not require precomputation and produces \emph{exact} geodesic distances, which are used for reference. 
The Heat Method (HM)~\cite{Crane2013a,Crane2017} computes geodesic distances by solving a heat equation and a Poisson equation on meshes, with a time complexity of $\mathcal{O}(N)$ for each GDQ after matrix factorization of the involved linear system.
The Discrete Geodesic Graphs (DGG)~\cite{Adikusuma2020} construct an undirected and sparse graph for computing discrete geodesic distances on triangle meshes and requires $\mathcal{O}(N)$ time for each GDQ.
The optimization-based methods, including the Euclidean Embedding Method (EEM)~\cite{Panozzo2013} and the Geodesic Embedding (GE)~\cite{Xia2021}, leverage costly cascaded non-linear optimization to compute the geodesic embedding of a mesh, with a time complexity of $\mathcal{O}(1)$ for each GDQ.
The precomputation time of FPGDC~\cite{shamai2018efficient} depends on its sample number $k$, with a time complexity of $\mathcal{O}(1)$ for each GDQ.
Our method produces the geodesic embedding by a forward evaluation of a GNN and has a time complexity of $\mathcal{O}(1)$ for each GDQ.
We compare our method with the above methods on the three groups of meshes using the code provided by the authors or publicly available on the web.
Previous methods are mainly developed with C++ and run on CPU, and it is non-trivial for us to reimplement them on GPU; 
we thus run them on top of a Windows computer with a Ryzen R9 5900HX CPU and 32GB of memory. We keep their default parameter settings unchanged during the evaluations unless specified.
We evaluate our method on a single RTX 3090 GPU with a batch size of 10. \looseness=-1

\begin{table}[t]
    \tablestyle{2.3pt}{1.1}
    \caption{Performances comparisons on geodesic distance queries. Algorithms specifically designed for GDQ are in bold.
    \emph{PC} represents the average time of precomputation for GQDs.
    \emph{GDQ} represents the average time of 1 million GDQs after any precomputation.
    \emph{MRE} represents the mean relative error.
    Compared with the state-of-the-art methods, including MMP~\cite{Mitchell1987}, HM~\cite{Crane2013a, Crane2017}, DGG~\cite{Adikusuma2020}, EEM~\cite{Panozzo2013}, FPGDC~\cite{shamai2018efficient} and GE~\cite{Xia2021}, our \textsc{GeGnn} achieves comparable accuracies while being significantly faster in terms of both precomputation and GDQs by orders of magnitude, as indicated in the \emph{Speedup} column with green color. 
    }
    \begin{tabular}{ccccccc}
    \toprule
    \multirow{2}{*}{Method} & \multirow{2}{*}{Complexity} & \multirow{2}{*}{Speedup} & \multirow{2}{*}{Metric} & ShapeNet-A & ShapeNet-B & Common   \\ 
    & & & & 5,057 & 5,120 & 7,202    \\ \midrule

    {MMP} & {$\mathcal{O}(N^2 \log N)$} & \hl{$1.7\! \times\! 10^6$} & GDQ  & 15.85h    & 18.00h   &  25.41h \\ 
                  \midrule
    \multirow{2}{*}{HM}     & \multirow{2}{*}{$\mathcal{O}(N)$}   & \hl{$6.7\! \times\! 10^4$} & GDQ             & 2345s & 2406s & 3506s \\ 
                   & & & MRE             & 2.41\% & 2.04\% & 1.46\% \\  \midrule
    \multirow{2}{*}{DGG}   &  \multirow{2}{*}{$\mathcal{O}(N)$} & \hl{$3.2\! \times\! 10^4$} & GDQ             & 1163s & 1157s & 1626s \\ 
                   & & & MRE             & 0.27\% & 0.27\% & 0.50\% \\ \midrule
                   & & \hl{$4.1\! \times\! 10^2$} & PC             & 45.02s    & 30.87s   & 41.21s \\ 
    \textbf{EEM}   & $\mathcal{O}(1)$  & \hl{$3.2\! \times\! 10^0$} & GDQ            & 0.134s    & 0.120s        & 0.136s \\ 
                   & & & MRE           & 8.11\%      & 8.78\% & 5.64\% \\ \midrule

        & & \hl{$7.1\! \times\! 10^1$} & PC            & 5.64s    & 6.28s   & 8.23s \\ 
    \textbf{FPGDC}   & $\mathcal{O}(1)$  & \hl{$6.2\! \times\! 10^1$} & GDQ            & 2.75s    & 2.37s        & 2.56s \\ 
                   & & & MRE           & 2.50\%      & 2.30\% & 2.30\% \\ \midrule

    \textbf{GE}               & $\mathcal{O}(1)$ & \hl{$3.2\! \times\! 10^3$}& PC              & 225.5s & 297.8s & 387.8s \\   
    
                   \midrule
                   & & & PC              & 0.089s & 0.091s & 0.104s \\ 
    \textbf{\textsc{GeGnn}}  & $\mathcal{O}(1)$ & &GDQ             & 0.042s  & 0.040s & 0.041s \\ 
                   & & & MRE             & 1.33\% & 2.55\% & 2.30\% \\
    \bottomrule
    \end{tabular}
    \label{tab:results}
\end{table}

\paragraph{Speed}
The key benefit of our method is its efficiency in terms of both precomputation and GDQs compared with the state-of-the-art methods.
The results are summarized in \cref{tab:results}.
Since MMP, HM, and DGG are targeted at computing single-source-all-destinations geodesic distances, our method is at least $\mathbf{3.2 \times 10^4}$ \emph{\textbf{times faster}} than them when computing GDQs.
Compared with EEM, GE and FPGDC, which all have $\mathcal{O}(1)$ complexity for GDQs, in terms of precomputation, our method is $\mathbf{3.2 \times 10^3}$ \emph{\textbf{times faster}} than GE, $\mathbf{4.1 \times 10^2}$ \emph{\textbf{times faster}} than EEM,
$\mathbf{71}$ \emph{\textbf{times faster}} than FPGDC with its parameter $N$ set to 400.
Here, the precomputation time is the time required to compute the geodesic embedding of a mesh, which is not applicable to MMP, HM, and SVG.
The precomputation of our \textsc{GeGnn} involves a single forward pass of the network, which is trivially parallelized on GPUs.
In contrast, the precomputation of GE, FPGDC and  EMM requires a costly non-linear optimization process or eigen decomposition, which is hard to take advantage of GPU parallelism without sophisticated optimization efforts.
Additionally, each GDQ of our method only involves a few matrix products, which are also highly optimized and parallelized on GPUs, whereas the GDQ of GE involves geodesic computation with the help of saddle vertex graphs.
We did not calculate the GDQ and MRE of GE in \cref{tab:results} since the authors only provided the code for precomputation.

\paragraph{Accuracy}
In \cref{tab:results}, MMP~\cite{Mitchell1987} computes the exact geodesic distances, which is used as the ground truth;  the other methods compute approximate geodesic distances.
We observe that the accuracy of FPGDC is severely affected by the sample number. 
With sample number 20 (the default value), its MRE is 12\%, much worse than the other methods.
Therefore, we set its sample number $k$ to $400$, with which it achieves an accuracy that is slightly worse than ours.
Note that FPGDC fails on over 5\% of samples in ShapeNet-A and ShapeNet-B; EEM also fails on around 2\% meshes.
We exclude those failed samples when computing their accuracies.
It can be seen that the accuracy of our \textsc{GeGnn} is significantly better than EEM~\cite{Panozzo2013}, and our \textsc{GeGnn} even outperforms the Heat Method~\cite{Crane2013a,Crane2017} on ShapeNet-A. 
However, our accuracy is slightly worse than DGG~\cite{Adikusuma2020}.
Nevertheless, we emphasize that it does not affect the applicability of our method in many graphics applications with a mean relative error of less than 3\%, including texture mapping, shape analysis as verified in \cref{subsec:application},  and many other applications as demonstrated in EEM~\cite{Panozzo2013}, of which the mean relative error is above 8\% on ShapeNet.

\begin{figure}[t]
    \centering
    \includegraphics[width=\linewidth]{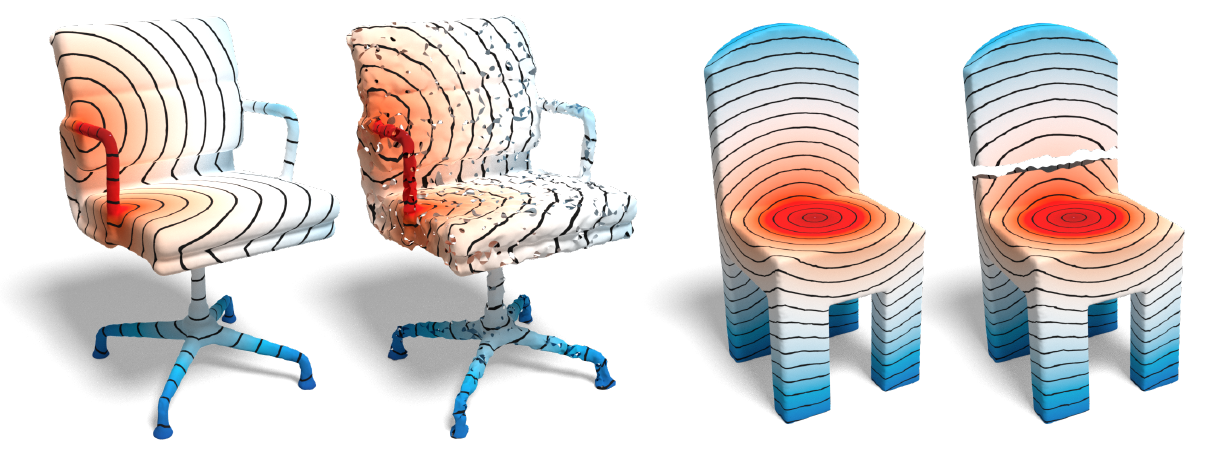}
    \caption{Robustness of our method. 
    Left: the result on an incomplete and noisy mesh. Right: the result on a  mesh containing two separated parts.
     }
    \label{fig:robust}
\end{figure}

\begin{figure*}[t]
    \includegraphics[width=0.95\linewidth]{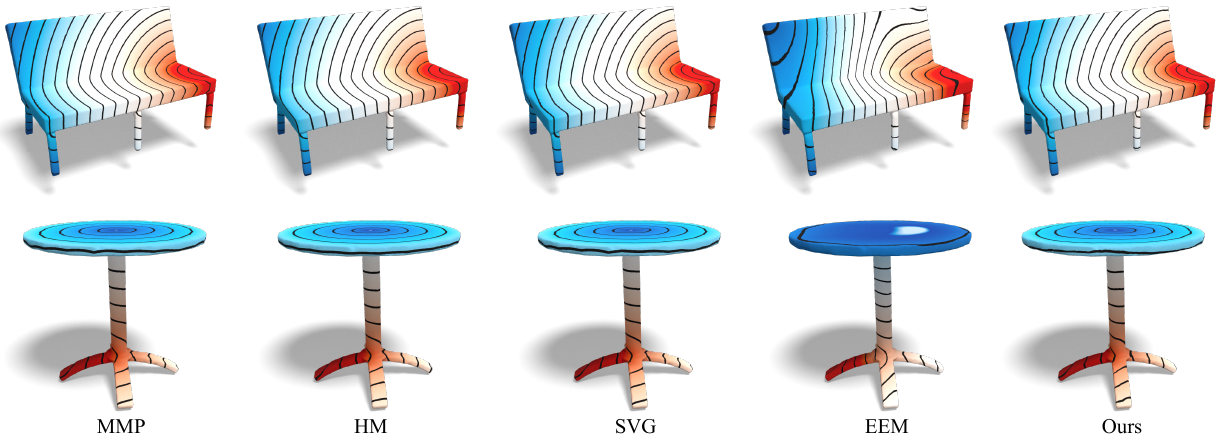}
    \caption{Visual Comparisons.
    We compared our results with the results of MMP~\cite{Mitchell1987}, DDG~\cite{Adikusuma2020}, EEM~\cite{Panozzo2013}, and HM~\cite{Crane2017}.
    All methods apart from EEM get visually pleasing results. 
    }
    \label{fig:compare}
\end{figure*}

\begin{figure*}[t]
    \includegraphics[width=\linewidth]{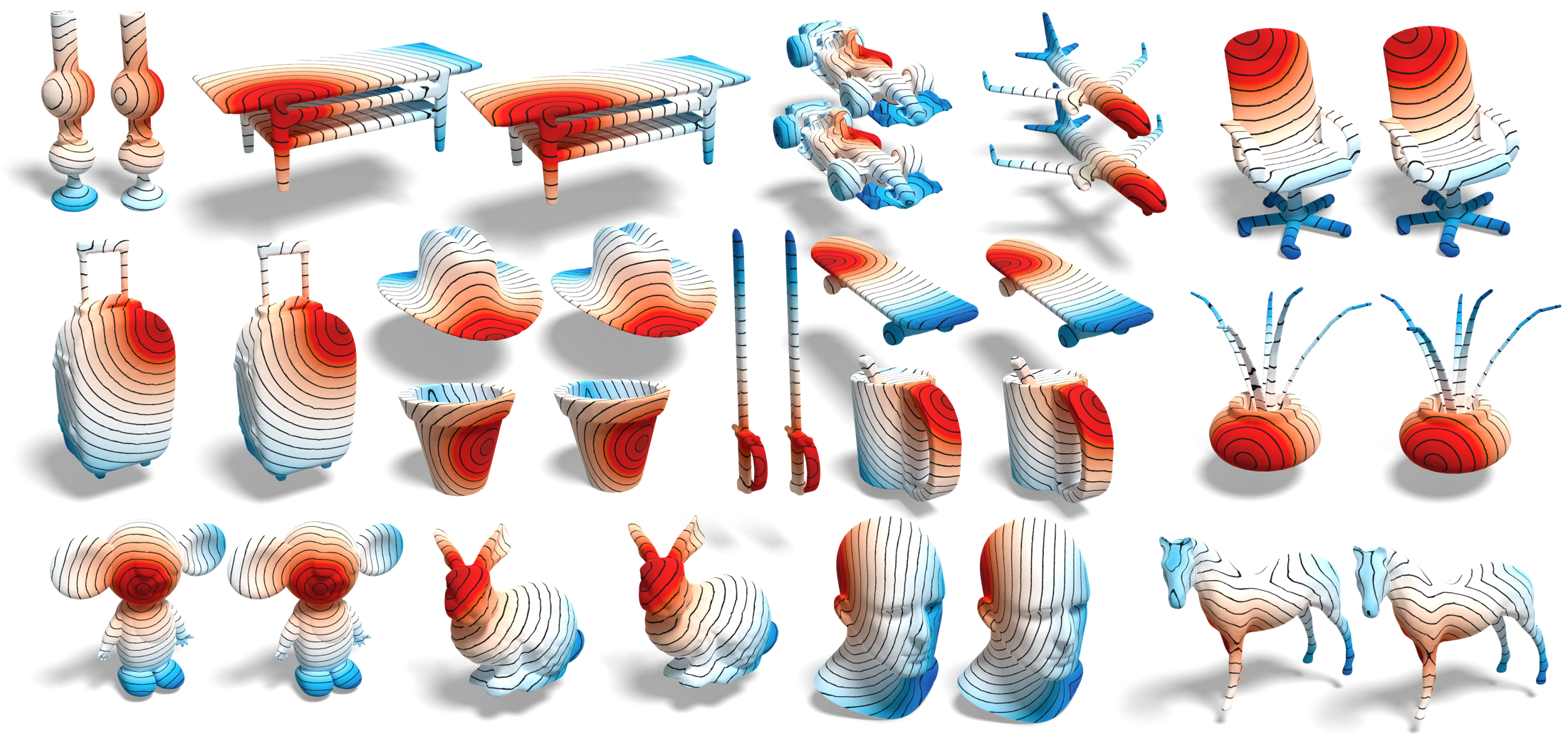}
    \caption{More results on meshes with complex geometric or topological structures.
    For each pair of meshes, our results are shown on the left or top side, and the corresponding ground truth results are shown on the right or bottom side.
    The meshes from top to bottom are selected from ShapeNet-A, ShapeNet-B, and Common, respectively.
    These results demonstrate the effectiveness and strong generalization ability of our method.
    }
    \label{fig:results}
\end{figure*}

\paragraph{Visual Results}
We compare the geodesic distances of our method with other methods in \cref{fig:compare}. 
And we show more geodesic distance fields generated by our method in \cref{fig:results} on various shapes from group ShapeNet-A, ShapeNet-B, and Common in each of the three rows, respectively.
These results demonstrate the effectiveness of our \textsc{GeGnn} in computing geodesic distances on meshes with different topologies and geometries, such as those with high genus, curved surfaces, and complex features. 
It is worth highlighting that our network is trained on ShapeNet-A, and it can generalize well to ShapeNet-B and Common, which demonstrates the strong generalization ability of our method.

\paragraph{Robustness}
After training, our \textsc{GeGnn} learns shape priors from the dataset, endowing it with strong robustness to incomplete or noisy meshes.
On the left of \cref{fig:robust}, we randomly remove 15\% of the triangles and add a Gaussian noise with a standard deviation of 0.06 on every vertex. Applying our method to the incomplete and noisy mesh results in an MRE of 2.76\%.
On the right of \cref{fig:robust}, we removed a continuous triangular area, separating the mesh into two parts. Traditional methods cannot handle this case, while our method can still generate a good result.

\begin{figure}[t]
    \centering
    \includegraphics[width=0.95\linewidth]{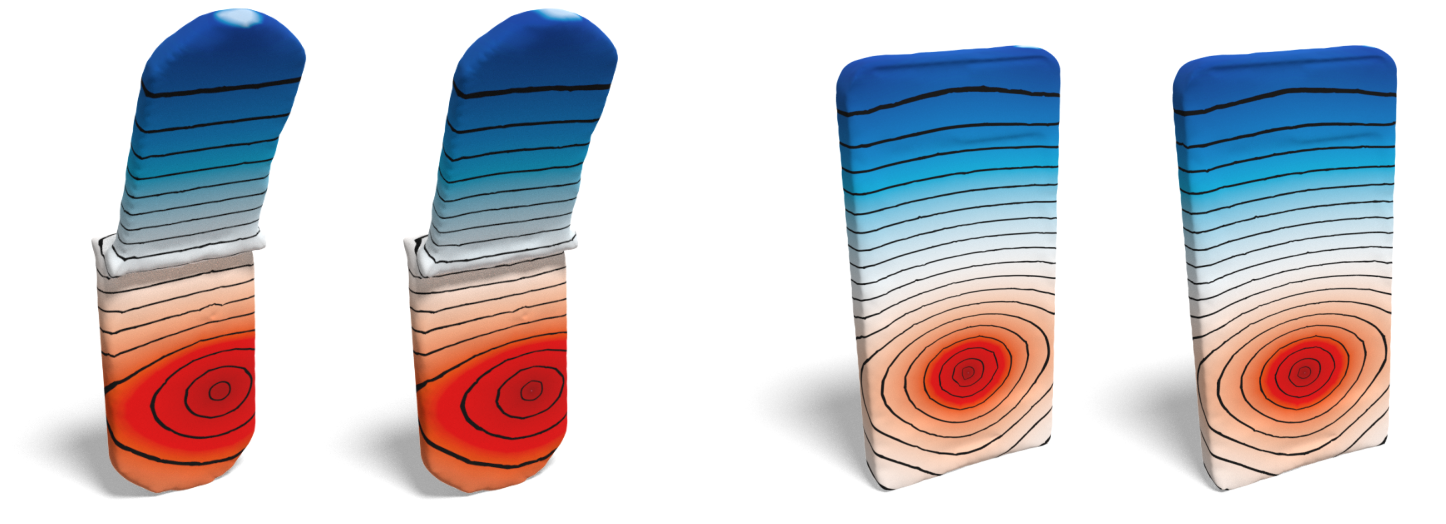}
    \caption{Train our network to predict the biharmonic distance. 
    In each pair of results, the exact biharmonic distance is on the left, and the predicted biharmonic distance is on the right. Our model predicts the biharmonic distance with high accuracy. }
    \label{fig:biharmonic}
\end{figure}

\paragraph{Biharmonic Distance}
In this experiment, we verify the versatility of our method on learning high-dimensional embeddings for the biharmonic distance~\cite{Lipman2010}.
We select 900 meshes from the training data and compute their ground-truth biharmonic distance.
Then, we train the network in a similar way as learning the geodesic embedding.
After training, we can also query the biharmonic distances between two arbitrary points, eliminating the requirements of estimating eigenvectors and solving linear systems in~\cite{Lipman2010}.
We show the predicted results in \cref{fig:biharmonic}, which are faithful to the ground truth.
Potentially, it is possible to apply our method to learn embeddings for other types of distances on manifolds, such as the diffusion distance~\cite{Coifman2005} and the earth mover's distance~\cite{Solomon2014a}.

\begin{figure}[t]
    \includegraphics[width=\linewidth]{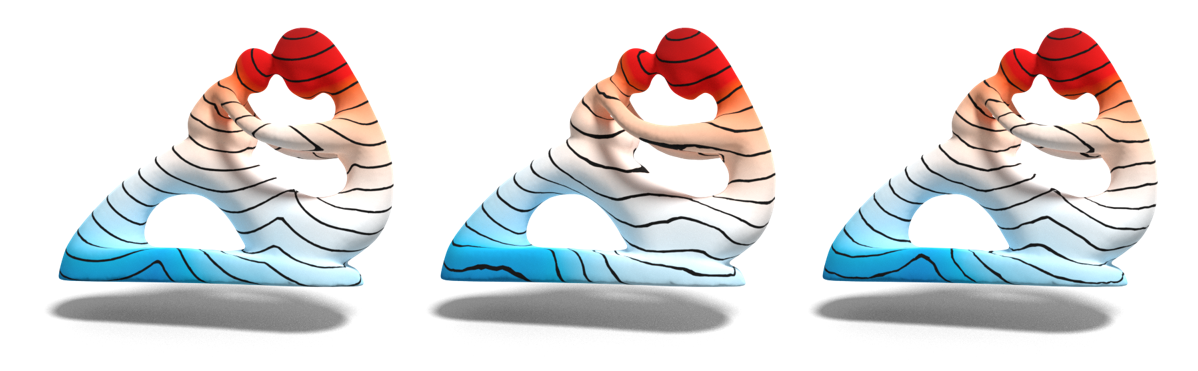}
    \caption{The effect of finetuning. Left: Exact geodesic distance. Middle: Predicted geodesic distance. Right: Predicted geodesic distance after 2,500 iterations of finetuning. Before finetuning, the MRE is 4.1\%, while after finetuning, the MRE is reduced to 1.0\%.}
    \label{fig:finetune}
\end{figure}

\paragraph{Finetune}
We also test the effect of finetuning
by additionally training the network with the ground-truth samples on the specified mesh to improve the performance of \textsc{GeGnn}. 
An example is shown in \cref{fig:finetune}. The mesh is not contained in ShapeNet and has a relatively complicated topology. Our network produces moderately accurate results initially. After finetuning, the predicted geodesic distance is much closer to the ground truth.

\begin{figure}[t]
    \includegraphics[width=\linewidth]{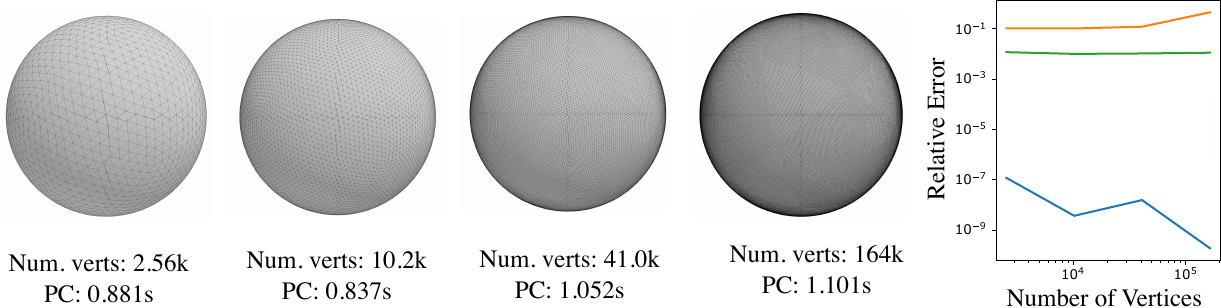}
    \caption{The convergence test. Left: The meshes with increasing number of vertices and their corresponding precomputation time. Right: The plot of mean (green), max (orange), and min (blue) relative error.}
    \label{fig:convergence}
\end{figure}

\paragraph{Scalability and Convergence}
Our network is fully convolutional and has good scalability with large meshes.
We firstly test our method on the ShapeNet-A dataset with meshes of different sizes. 
Specifically, we apply the Loop subdivision~\cite{Loop1987} on ShapeNet-A to get two sets of meshes with an average number of vertices of 18,708 and 23,681, respectively. Then, we test the speed and accuracy of our method on these meshes. The results are shown in \cref{tab:scalability}.
Although our network is trained on ShapeNet with an average number of vertices of 5,057, it can efficiently and accurately deal with mesh with many more vertices.
To verify the ability of our method on even larger meshes, we conduct experiments on subdivided spheres following \cite{Surazhsky2005}. 
We report the accuracy and running time in \cref{fig:convergence}.
The mean, maximum, and minimum relative errors on meshes are almost constant on spheres with a wide range of resolutions.
And the running time only increases slightly when the resolution increases since the network runs in parallel on GPUs.
The results demonstrate that our algorithm exhibits good convergence in terms of mesh resolution.

\begin{table}[t]
     \tablestyle{16pt}{1.1}
    \caption{The scalability test about the number of vertices. The ``Ratio'' represents the ratio between the average number of vertices of the test meshes and the average number of vertices in our training dataset.}
     \label{tab:scalability}
     \begin{tabular}{lccc}
     \toprule
         Avg. Vert. & 23682 & 18708 & 5202 \\  \midrule
         Ratio & 4.7 & 3.7 & 1.0 \\ 
         PC & 0.1395s & 0.1311s & 0.089s \\ 
         MRE & 2.00\% & 1.88\% & 1.33\% \\
              \bottomrule
     \end{tabular}
 \end{table}

\subsection{Ablation Studies}
In this section, we study and discuss the effectiveness of key designs in our method, including the graph convolution module, the graph pooling module, and the decoding function. 
We compare the MRE on the whole testing set, including 4,732 meshes.

\paragraph{Graph Convolution}
The two key designs of our graph convolution are the local geometric features and the \emph{max} aggregator in \cref{eq:conv}.
To verify the effectiveness of these designs, we conduct a set of experiments to compare the performance; the mean relative errors are listed as follows:
\begin{center}
    \vspace{2pt}
    \tablestyle{8pt}{1.1}
    \begin{tabular}{c|ccccc}
          & GeoConv & mean & sum & w/o dist  & w/o rel. pos.  \\ \midrule
     MRE  & \textbf{1.57}\%  & 2.53\% & 2.53\% & 1.67\%  & 1.84\% \\
    \end{tabular}
    \vspace{2pt}
\end{center}
After replacing the max aggregator with \emph{mean} or \emph{sum}, as shown in the third and fourth columns of the table, the performance drops by 0.96\%.
After removing the pairwise edge length (w/o dist) and the relative position (w/o rel. pos.), the performance drops by 0.27\% and 0.67\%, respectively.

We also compare our GeoConv with other graph convolutions, including GraphSAGE~\cite{Hamilton2017}, GCN~\cite{Kipf2017}, GATv2~\cite{Brody2022}, and EdgeConv~\cite{Wang2018}.
We keep all the other settings the same as our method, except that we halve the batch size GATv2 and EdgeConv; otherwise, the models will run out of GPU memory.
The results are listed as follows:
\begin{center}
    \vspace{2pt}
    \tablestyle{7pt}{1.1}
    \begin{tabular}{c|ccccc}
          & GeoConv  & GraphSAGE & GCN & GATv2 & EdgeConv   \\ \midrule
     MRE  & \textbf{1.57}\%  & 2.84\%  & 3.25\% & 2.34\%  & 1.62\% \\
    \end{tabular}
    \vspace{2pt}
\end{center}
It can be seen that although our GeoConv in \cref{eq:conv} only contains two trainable weights, it still outperforms all the other graph convolutions by a large margin in the task of geodesic embedding.
Our GeoConv is easy to implement and can be implemented within 15 lines of code with the Torch Geometric library~\cite{Fey2019}, which we expect to be useful for other geometric learning tasks. 
\looseness=-1

\paragraph{Graph Pooling}
In this experiment, we replace our GeoPool with naive grid-based pooling to verify the efficacy of our graph pooling.
We did not compare the pooling strategy in MeshCNN~\cite{Hanocka2019} and SubdivConv~\cite{Hu2022} since they require sequential operations on CPUs and are time-consuming for our task. 
The MRE of grid-based pooling is 2.04\%, which is 0.47\% worse than our GeoPool.

\paragraph{Decoding Function}
In this experiment, we verify the superiority of our MLP-based function over conventional Euclidean distance~\cite{Panozzo2013,Xia2021} when decoding the geodesic distance from the embedding vectors.
After replacing our MLP with Euclidean distance, the MRE is 3.03\%, decreasing by 1.46\%.

\subsection{Applications} \label{subsec:application}

In this section, we demonstrate several interesting applications with our constant-time GDQs.

\begin{figure}[t]
    \centering
    \includegraphics[width=\linewidth]{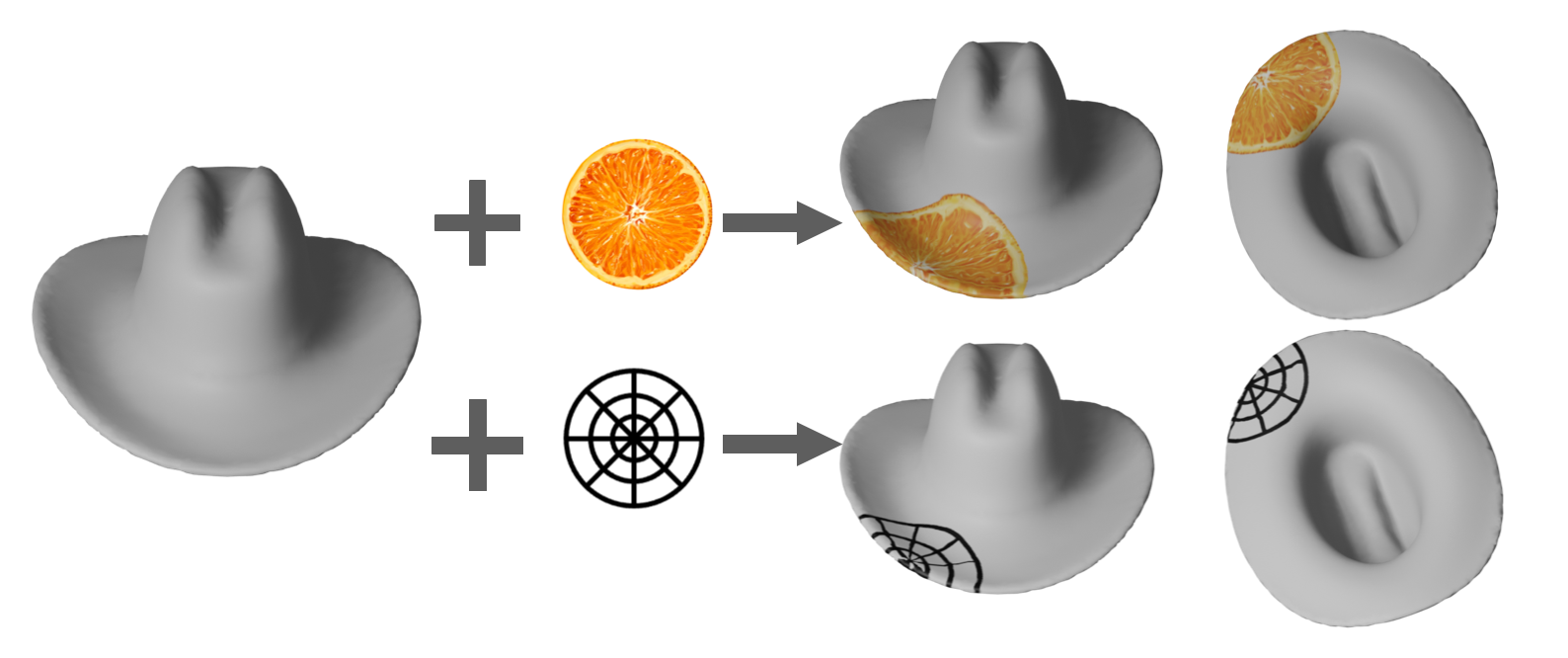}
    \caption{Texture Mapping.
    We compute the exponential map with our constant-time GDQs (below) and map the texture from the polar coordinate system to the mesh (top). 
    }
    \label{fig:texture}
\end{figure}

\paragraph{Texture Mapping}
We create a local geodesic polar coordinate system around a specified point on the mesh, which is known as the logarithmic map in differential geometry, also called the logarithmic map.
Then, we can map the texture from the polar coordinate system to the mesh.
We follow \cite{Xin2012,Panozzo2013} to compute the exponential map with constant-time GDQs, which has proven to be more efficient and more accurate than \cite{Schmidt2006}.
The results are shown in \cref{fig:texture}, where the texture is smoothly mapped to curved meshes with the computed exponential map, demonstrating that our algorithm is robust even on the edge of the surface.

\begin{figure}[t]
    \centering
    \includegraphics[width=\linewidth]{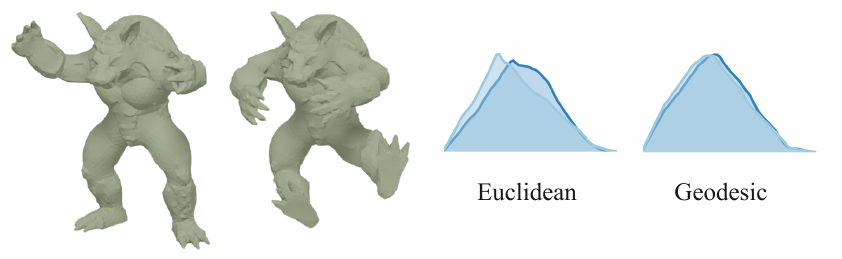}
    \caption{Shape distributions. We use Blender to deform the mesh of the armadillo and compute the shape distributions with geodesic distances and Euclidean distances as shape functions, respectively.
    The shape distributions computed with geodesic distances are more invariant to the articulated deformation of the mesh than Euclidean distances.
    }
    
    \label{fig:matching}
\end{figure}

\paragraph{Shape Matching}
Shape distributions~\cite{Osada2002} are widely used in shape matching and retrieval, which is based on shape histograms of pairwise distances between many point pairs on the mesh using a specific shape function.
In \cite{Osada2002}, the shape function is defined as the Euclidean distance or angle difference between two points.
We follow \cite{Martinek2012} to use the geodesic distances as the shape function. Our constant-time GDQs allow for the extremely efficient computation of geodesic distances between randomly selected surface points.
The results are shown in \cref{fig:matching}; we can see that the shape distributions computed with geodesic distances are invariant to the articulated deformation of the mesh, while the shape distributions computed with Euclidean distances are not.

\begin{wrapfigure}[7]{r}{0.25\linewidth}
    \centering
    \vspace{-12pt}
    \hspace{-20pt}
    \includegraphics[trim=10 0 10 0,clip,width=1.1\linewidth]{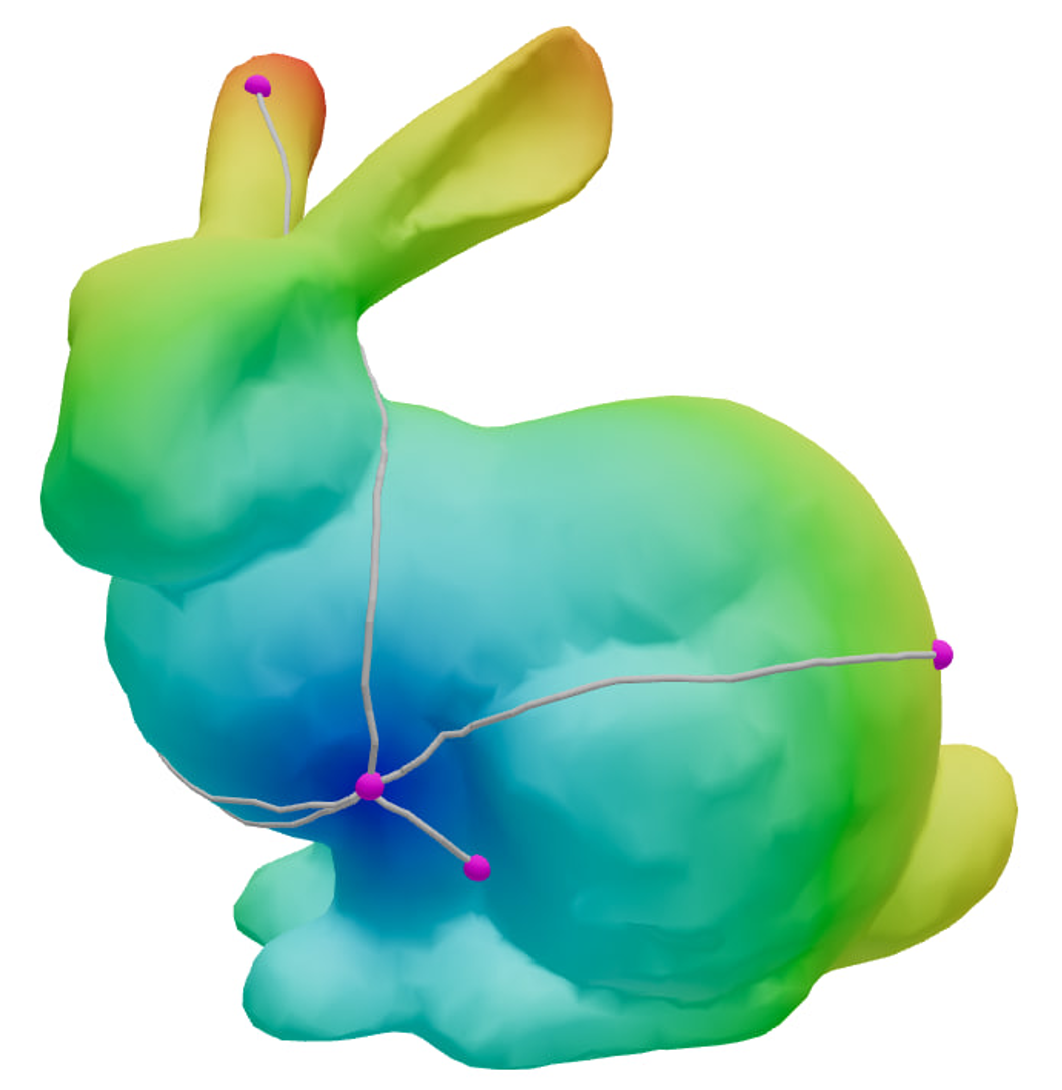}
\end{wrapfigure}
\paragraph{Geodesic Path}
We can also efficiently compute the geodesic path by tracing the gradient of the geodesic distance field from the target point to the source point~\cite{Xin2012,Kimmel1998}.
The gradient of the geodesic distance field can be computed in constant time on each triangle with our method, therefore the geodesic path can be computed in $\mathcal{O}(k)$ time, where $k$ is the number of edges crossed by the path.
The results are shown on the right.
For visualization, the geodesic distance field is also drawn on the mesh. \looseness=-1

%% file: src/conclusion.tex
\section {Discussion and limitations} \label{sec:discussion}

\begin{figure*}[t]
    \centering
    \includegraphics[width=0.9\linewidth]{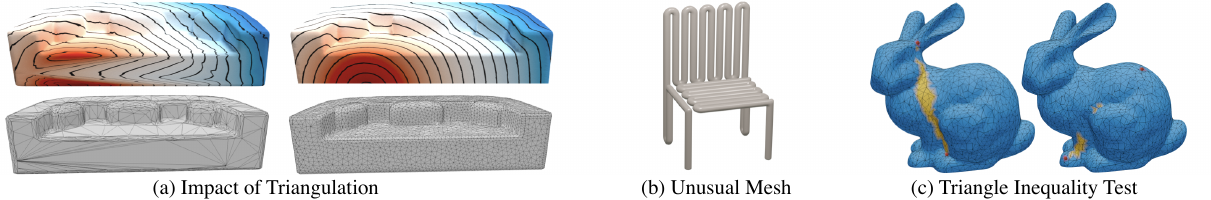}
    \caption{The limitations of our method. 
    (a) Our method cannot handle the extremely anisotropic mesh; the result on a regular mesh is much better, as shown on the right.
    (b) Our method cannot handle shapes with unusual geometry.
    (c) Our method cannot guarantee the triangle inequality for the predicted geodesic distances.
    For fixed vertices marked in red, all vertices for which triangle inequality is not satisfied are marked in orange.
    }
    \label{fig:limitation_overall}
\end{figure*}

In this section, we discuss our method's limitations and possible improvements.

\paragraph{Triangulations}
Although our method exhibits good robustness against different triangulations, it cannot work well with extremely anisotropic meshes; an example is shown in \cref{fig:limitation_overall}-(a).
However, this issue could be easily alleviated by making the mesh regular through remeshing.

\paragraph{Unusual meshes}
Our method may struggle to process meshes that have unusual topology and geometry. The chair in \cref{fig:limitation_overall}-(b) is constructed by bending a long and thin tube multiple times; such geometry is rare in the training dataset. 
As a consequence, our method produces an MRE of about 25\%, significantly higher than the average error throughout the test set. 
In order to address this issue, we could potentially expand and diversify our training data while also enlarging our networks to enhance the overall generalizability and robustness.
\looseness=-1

\paragraph{Triangle Inequality and Error Bound}
The geodesic distance satisfies the triangle inequality.
However, our \textsc{GeGnn} only predicts the approximate geodesic distances and cannot strictly pass the triangle inequality test.
We follow~\cite{Solomon2014a} to illustrate this issue in~\cref{fig:limitation_overall} (c).
Specifically, given two fixed vertices $p$ and $q$, we identify and mark all vertices $x$ for which triangle inequality $d(p,q) \geq d(p,x) + d(q,x)$ are not satisfied.
Our results are visually on par with the results of~\cite{Crane2013}.
As a learning-based method, our \textsc{GeGnn} cannot offer theoretical guarantees of error bound, either.
It would be interesting to incorporate additional geometric priors to address this limitation.

\paragraph{Rotation and Translation Invariance}
Although the \emph{GeoConv} in \cref{eq:conv} is translation-invariant, the network is not due to the grid sampling in pooling.
Our network is not rotation invariant, either.
It is possible to leverage network architectures that possess invariant properties~\cite{deng2021vector} to increase the performance further.

\section{Conclusion} \label{sec:conclusion}

We propose to learn the geodesic embedding with graph neural networks, which enables constant time complexity for geodesic distance queries on discrete surfaces with arbitrary topology and geometry.
We design a novel graph neural network to predict the vertex-wise geodesic embedding and leverage a lightweight MLP to decode the geodesic distance from the embeddings.
The key technical contributions of our method include a novel graph convolution, graph pooling, and the design of the MLP decoder.
We verify the efficiency, effectiveness, robustness, and generalizability of our method on ShapeNet and a variety of out-of-distribution meshes.
We expect our work can inspire more learning-based methods for geodesic distance or general geometry problems. 
Several  future works are discussed as follows.

\paragraph{General Geometric Distances}
In this paper, we mainly use \textsc{GeGnn} to learn the geodesic embedding and verify the feasibility in learning an embedding for biharmonic distances.
In the future, it is interesting to explore the possibility of learning other geometric distances and, furthermore, training a single general model for all geometric distances on meshes.

\paragraph{Geometry Optimization}
The process of computing geodesic embedding is essentially a geometry optimization problem on meshes.
In the future, we plan to use graph neural networks to solve other geometry optimization problems, such as shape deformation, remeshing, and parameterization.

\paragraph{Transformers on Meshes}
Transformers have been prevailing in natural language processing and computer vision.
It is interesting to use Transformers to meshes to replace the graph neural networks in our method for better performance in the future.

%% file: GnnDist.bbl

\begin{thebibliography}{78}


\ifx \showCODEN    \undefined \def \showCODEN     #1{\unskip}     \fi
\ifx \showDOI      \undefined \def \showDOI       #1{#1}\fi
\ifx \showISBNx    \undefined \def \showISBNx     #1{\unskip}     \fi
\ifx \showISBNxiii \undefined \def \showISBNxiii  #1{\unskip}     \fi
\ifx \showISSN     \undefined \def \showISSN      #1{\unskip}     \fi
\ifx \showLCCN     \undefined \def \showLCCN      #1{\unskip}     \fi
\ifx \shownote     \undefined \def \shownote      #1{#1}          \fi
\ifx \showarticletitle \undefined \def \showarticletitle #1{#1}   \fi
\ifx \showURL      \undefined \def \showURL       {\relax}        \fi
\providecommand\bibfield[2]{#2}
\providecommand\bibinfo[2]{#2}
\providecommand\natexlab[1]{#1}
\providecommand\showeprint[2][]{arXiv:#2}

\bibitem[Adikusuma et~al\mbox{.}(2020)]%
        {Adikusuma2020}
\bibfield{author}{\bibinfo{person}{Yohanes~Yudhi Adikusuma},
  \bibinfo{person}{Zheng Fang}, {and} \bibinfo{person}{Ying He}.}
  \bibinfo{year}{2020}\natexlab{}.
\newblock \showarticletitle{Fast construction of discrete geodesic graphs}.
\newblock \bibinfo{journal}{\emph{ACM Trans. Graph.}} \bibinfo{volume}{39},
  \bibinfo{number}{2} (\bibinfo{year}{2020}).
\newblock


\bibitem[Bendels and Klein(2003)]%
        {Bendels2003}
\bibfield{author}{\bibinfo{person}{G.~H. Bendels} {and} \bibinfo{person}{R.
  Klein}.} \bibinfo{year}{2003}\natexlab{}.
\newblock \showarticletitle{Mesh Forging: Editing of {3D}-Meshes Using
  Implicitly Defined Occluders}. In \bibinfo{booktitle}{\emph{SGP}}.
\newblock


\bibitem[Bhat et~al\mbox{.}(2004)]%
        {Pravin2004}
\bibfield{author}{\bibinfo{person}{Pravin Bhat}, \bibinfo{person}{Stephen
  Ingram}, {and} \bibinfo{person}{Greg Turk}.} \bibinfo{year}{2004}\natexlab{}.
\newblock \showarticletitle{Geometric texture synthesis by example}. In
  \bibinfo{booktitle}{\emph{Symp. Geom. Proc.}}
\newblock


\bibitem[Brody et~al\mbox{.}(2022)]%
        {Brody2022}
\bibfield{author}{\bibinfo{person}{Shaked Brody}, \bibinfo{person}{Uri Alon},
  {and} \bibinfo{person}{Eran Yahav}.} \bibinfo{year}{2022}\natexlab{}.
\newblock \showarticletitle{How Attentive are Graph Attention Networks?}. In
  \bibinfo{booktitle}{\emph{ICLR}}.
\newblock


\bibitem[Carroll and Arabie(1998)]%
        {Carroll1998}
\bibfield{author}{\bibinfo{person}{J.~Douglas Carroll} {and}
  \bibinfo{person}{Phipps Arabie}.} \bibinfo{year}{1998}\natexlab{}.
\newblock \showarticletitle{Multidimensional scaling}.
\newblock \bibinfo{journal}{\emph{Measurement, judgment and decision making}}
  (\bibinfo{year}{1998}).
\newblock


\bibitem[Chang et~al\mbox{.}(2015)]%
        {Chang2015}
\bibfield{author}{\bibinfo{person}{Angel~X. Chang}, \bibinfo{person}{Thomas
  Funkhouser}, \bibinfo{person}{Leonidas~J. Guibas}, \bibinfo{person}{Pat
  Hanrahan}, \bibinfo{person}{Qixing Huang}, \bibinfo{person}{Zimo Li},
  \bibinfo{person}{Silvio Savarese}, \bibinfo{person}{Manolis Savva},
  \bibinfo{person}{Shuran Song}, \bibinfo{person}{Hao Su},
  \bibinfo{person}{Jianxiong Xiao}, \bibinfo{person}{Li Yi}, {and}
  \bibinfo{person}{Fisher Yu}.} \bibinfo{year}{2015}\natexlab{}.
\newblock \showarticletitle{{ShapeNet}: An information-rich {3D} model
  repository}.
\newblock \bibinfo{journal}{\emph{arXiv preprint arXiv:1512.03012}}
  (\bibinfo{year}{2015}).
\newblock


\bibitem[Chen and Han(1990)]%
        {Chen1990}
\bibfield{author}{\bibinfo{person}{Jindong Chen} {and} \bibinfo{person}{Yijie
  Han}.} \bibinfo{year}{1990}\natexlab{}.
\newblock \showarticletitle{Shortest paths on a polyhedron}. In
  \bibinfo{booktitle}{\emph{Proceedings of the sixth annual symposium on
  Computational geometry}}.
\newblock


\bibitem[Coifman et~al\mbox{.}(2005)]%
        {Coifman2005}
\bibfield{author}{\bibinfo{person}{Ronald~R Coifman}, \bibinfo{person}{Stephane
  Lafon}, \bibinfo{person}{Ann~B Lee}, \bibinfo{person}{Mauro Maggioni},
  \bibinfo{person}{Boaz Nadler}, \bibinfo{person}{Frederick Warner}, {and}
  \bibinfo{person}{Steven~W Zucker}.} \bibinfo{year}{2005}\natexlab{}.
\newblock \showarticletitle{Geometric diffusions as a tool for harmonic
  analysis and structure definition of data: Diffusion maps}.
\newblock \bibinfo{journal}{\emph{Proceedings of the national academy of
  sciences}} \bibinfo{volume}{102}, \bibinfo{number}{21}
  (\bibinfo{year}{2005}).
\newblock


\bibitem[Crane et~al\mbox{.}(2013a)]%
        {Crane2013}
\bibfield{author}{\bibinfo{person}{Keenan Crane}, \bibinfo{person}{Fernando de
  Goes}, \bibinfo{person}{Mathieu Desbrun}, {and} \bibinfo{person}{Peter
  Schr\"{o}der}.} \bibinfo{year}{2013}\natexlab{a}.
\newblock \showarticletitle{Digital geometry processing with discrete exterior
  calculus}. In \bibinfo{booktitle}{\emph{ACM SIGGRAPH 2013 courses}}.
\newblock


\bibitem[Crane et~al\mbox{.}(2020)]%
        {Crane2020}
\bibfield{author}{\bibinfo{person}{Keenan Crane}, \bibinfo{person}{Marco
  Livesu}, \bibinfo{person}{Enrico Puppo}, {and} \bibinfo{person}{Yipeng Qin}.}
  \bibinfo{year}{2020}\natexlab{}.
\newblock \showarticletitle{A survey of algorithms for geodesic paths and
  distances}.
\newblock \bibinfo{journal}{\emph{arXiv preprint arXiv:2007.10430}}
  (\bibinfo{year}{2020}).
\newblock


\bibitem[Crane et~al\mbox{.}(2013b)]%
        {Crane2013a}
\bibfield{author}{\bibinfo{person}{Keenan Crane}, \bibinfo{person}{Clarisse
  Weischedel}, {and} \bibinfo{person}{Max Wardetzky}.}
  \bibinfo{year}{2013}\natexlab{b}.
\newblock \showarticletitle{Geodesics in heat: A new approach to computing
  distance based on heat flow}.
\newblock \bibinfo{journal}{\emph{ACM Trans. Graph.}} \bibinfo{volume}{32},
  \bibinfo{number}{5} (\bibinfo{year}{2013}).
\newblock


\bibitem[Crane et~al\mbox{.}(2017)]%
        {Crane2017}
\bibfield{author}{\bibinfo{person}{Keenan Crane}, \bibinfo{person}{Clarisse
  Weischedel}, {and} \bibinfo{person}{Max Wardetzky}.}
  \bibinfo{year}{2017}\natexlab{}.
\newblock \showarticletitle{The Heat Method for Distance Computation}.
\newblock \bibinfo{journal}{\emph{Commun. ACM}} \bibinfo{volume}{60},
  \bibinfo{number}{11} (\bibinfo{year}{2017}).
\newblock


\bibitem[Deng et~al\mbox{.}(2021)]%
        {deng2021vector}
\bibfield{author}{\bibinfo{person}{Congyue Deng}, \bibinfo{person}{Or Litany},
  \bibinfo{person}{Yueqi Duan}, \bibinfo{person}{Adrien Poulenard},
  \bibinfo{person}{Andrea Tagliasacchi}, {and} \bibinfo{person}{Leonidas~J
  Guibas}.} \bibinfo{year}{2021}\natexlab{}.
\newblock \showarticletitle{Vector neurons: A general framework for so
  (3)-equivariant networks}. In \bibinfo{booktitle}{\emph{CVPR}}.
\newblock


\bibitem[Fey and Lenssen(2019)]%
        {Fey2019}
\bibfield{author}{\bibinfo{person}{Matthias Fey} {and}
  \bibinfo{person}{Jan~Eric Lenssen}.} \bibinfo{year}{2019}\natexlab{}.
\newblock \showarticletitle{Fast Graph Representation Learning with {PyTorch
  Geometric}}. In \bibinfo{booktitle}{\emph{ICLR Workshop}}.
\newblock


\bibitem[Fey et~al\mbox{.}(2018)]%
        {Fey2018}
\bibfield{author}{\bibinfo{person}{Matthias Fey}, \bibinfo{person}{Jan~Eric
  Lenssen}, \bibinfo{person}{Frank Weichert}, {and} \bibinfo{person}{Heinrich
  M{\"u}ller}.} \bibinfo{year}{2018}\natexlab{}.
\newblock \showarticletitle{{SplineCNN}: Fast Geometric Deep Learning with
  Continuous {B}-Spline Kernels}. In \bibinfo{booktitle}{\emph{CVPR}}.
\newblock


\bibitem[Fouss et~al\mbox{.}(2007)]%
        {Fouss2007}
\bibfield{author}{\bibinfo{person}{Francois Fouss}, \bibinfo{person}{Alain
  Pirotte}, \bibinfo{person}{Jean-Michel Renders}, {and} \bibinfo{person}{Marco
  Saerens}.} \bibinfo{year}{2007}\natexlab{}.
\newblock \showarticletitle{Random-walk computation of similarities between
  nodes of a graph with application to collaborative recommendation}.
\newblock \bibinfo{journal}{\emph{IEEE Transactions on knowledge and data
  engineering}} \bibinfo{volume}{19}, \bibinfo{number}{3}
  (\bibinfo{year}{2007}).
\newblock


\bibitem[Garland and Heckbert(1997)]%
        {Garland1997}
\bibfield{author}{\bibinfo{person}{Michael Garland} {and}
  \bibinfo{person}{Paul~S Heckbert}.} \bibinfo{year}{1997}\natexlab{}.
\newblock \showarticletitle{Surface simplification using quadric error
  metrics}. In \bibinfo{booktitle}{\emph{SIGGRAPH}}.
\newblock


\bibitem[Gilmer et~al\mbox{.}(2017)]%
        {Gilmer2017}
\bibfield{author}{\bibinfo{person}{Justin Gilmer}, \bibinfo{person}{Samuel~S
  Schoenholz}, \bibinfo{person}{Patrick~F Riley}, \bibinfo{person}{Oriol
  Vinyals}, {and} \bibinfo{person}{George~E Dahl}.}
  \bibinfo{year}{2017}\natexlab{}.
\newblock \showarticletitle{Neural message passing for quantum chemistry}. In
  \bibinfo{booktitle}{\emph{ICML}}.
\newblock


\bibitem[Gotsman and Hormann(2022)]%
        {Gotsman2022}
\bibfield{author}{\bibinfo{person}{Craig Gotsman} {and} \bibinfo{person}{Kai
  Hormann}.} \bibinfo{year}{2022}\natexlab{}.
\newblock \showarticletitle{Compressing Geodesic Information for Fast
  Point-to-Point Geodesic Distance Queries}. In
  \bibinfo{booktitle}{\emph{SIGGRAPH Asia}}.
\newblock


\bibitem[Hamilton et~al\mbox{.}(2017)]%
        {Hamilton2017}
\bibfield{author}{\bibinfo{person}{Will Hamilton}, \bibinfo{person}{Zhitao
  Ying}, {and} \bibinfo{person}{Jure Leskovec}.}
  \bibinfo{year}{2017}\natexlab{}.
\newblock \showarticletitle{Inductive Representation Learning on Large Graphs}.
  In \bibinfo{booktitle}{\emph{NeurIPS}}.
\newblock


\bibitem[Hanocka et~al\mbox{.}(2019)]%
        {Hanocka2019}
\bibfield{author}{\bibinfo{person}{Rana Hanocka}, \bibinfo{person}{Amir Hertz},
  \bibinfo{person}{Noa Fish}, \bibinfo{person}{Raja Giryes},
  \bibinfo{person}{Shachar Fleishman}, {and} \bibinfo{person}{Daniel
  Cohen-Or}.} \bibinfo{year}{2019}\natexlab{}.
\newblock \showarticletitle{{MeshCNN}: A network with an edge}.
\newblock \bibinfo{journal}{\emph{ACM Trans. Graph. (SIGGRAPH)}}
  \bibinfo{volume}{38}, \bibinfo{number}{4} (\bibinfo{year}{2019}).
\newblock


\bibitem[Hanocka et~al\mbox{.}(2020)]%
        {Hanocka2020}
\bibfield{author}{\bibinfo{person}{Rana Hanocka}, \bibinfo{person}{Gal Metzer},
  \bibinfo{person}{Raja Giryes}, {and} \bibinfo{person}{Daniel Cohen-Or}.}
  \bibinfo{year}{2020}\natexlab{}.
\newblock \showarticletitle{{Point2Mesh}: A Self-Prior for Deformable Meshes}.
\newblock \bibinfo{journal}{\emph{ACM Trans. Graph. (SIGGRAPH)}}
  \bibinfo{volume}{39}, \bibinfo{number}{4} (\bibinfo{year}{2020}).
\newblock


\bibitem[He et~al\mbox{.}(2016)]%
        {He2016}
\bibfield{author}{\bibinfo{person}{Kaiming He}, \bibinfo{person}{Xiangyu
  Zhang}, \bibinfo{person}{Shaoqing Ren}, {and} \bibinfo{person}{Jian Sun}.}
  \bibinfo{year}{2016}\natexlab{}.
\newblock \showarticletitle{Deep residual learning for image recognition}. In
  \bibinfo{booktitle}{\emph{CVPR}}.
\newblock


\bibitem[Hoppe et~al\mbox{.}(1993)]%
        {Hoppe1993}
\bibfield{author}{\bibinfo{person}{Hugues Hoppe}, \bibinfo{person}{Tony
  DeRose}, \bibinfo{person}{Tom Duchamp}, \bibinfo{person}{John McDonald},
  {and} \bibinfo{person}{Werner Stuetzle}.} \bibinfo{year}{1993}\natexlab{}.
\newblock \showarticletitle{Mesh optimization}. In
  \bibinfo{booktitle}{\emph{SIGGRAPH}}.
\newblock


\bibitem[Hu et~al\mbox{.}(2020)]%
        {Hu2019}
\bibfield{author}{\bibinfo{person}{Qingyong Hu}, \bibinfo{person}{Bo Yang},
  \bibinfo{person}{Linhai Xie}, \bibinfo{person}{Stefano Rosa},
  \bibinfo{person}{Yulan Guo}, \bibinfo{person}{Zhihua Wang},
  \bibinfo{person}{Niki Trigoni}, {and} \bibinfo{person}{Andrew Markham}.}
  \bibinfo{year}{2020}\natexlab{}.
\newblock \showarticletitle{{RandLA-Net}: Efficient Semantic Segmentation of
  Large-Scale Point Clouds}.
\newblock \bibinfo{journal}{\emph{CVPR}}.
\newblock


\bibitem[Hu et~al\mbox{.}(2022)]%
        {Hu2022}
\bibfield{author}{\bibinfo{person}{Shi-Min Hu}, \bibinfo{person}{Zheng-Ning
  Liu}, \bibinfo{person}{Meng-Hao Guo}, \bibinfo{person}{Jun-Xiong Cai},
  \bibinfo{person}{Jiahui Huang}, \bibinfo{person}{Tai-Jiang Mu}, {and}
  \bibinfo{person}{Ralph~R Martin}.} \bibinfo{year}{2022}\natexlab{}.
\newblock \showarticletitle{Subdivision-based mesh convolution networks}.
\newblock \bibinfo{journal}{\emph{ACM Trans. Graph.}} \bibinfo{volume}{41},
  \bibinfo{number}{3} (\bibinfo{year}{2022}).
\newblock


\bibitem[Kimmel and Sethian(1998)]%
        {Kimmel1998}
\bibfield{author}{\bibinfo{person}{Ron Kimmel} {and} \bibinfo{person}{James~A.
  Sethian}.} \bibinfo{year}{1998}\natexlab{}.
\newblock \showarticletitle{Computing geodesic paths on manifolds}.
\newblock \bibinfo{journal}{\emph{Proceedings of the National Academy of
  Science}} \bibinfo{volume}{95}, \bibinfo{number}{15} (\bibinfo{year}{1998}).
\newblock


\bibitem[Kipf and Welling(2017)]%
        {Kipf2017}
\bibfield{author}{\bibinfo{person}{Thomas~N Kipf} {and} \bibinfo{person}{Max
  Welling}.} \bibinfo{year}{2017}\natexlab{}.
\newblock \showarticletitle{Semi-supervised classification with graph
  convolutional networks}. In \bibinfo{booktitle}{\emph{ICLR}}.
\newblock


\bibitem[Li et~al\mbox{.}(2018)]%
        {Li2018}
\bibfield{author}{\bibinfo{person}{Yangyan Li}, \bibinfo{person}{Rui Bu},
  \bibinfo{person}{Mingchao Sun}, \bibinfo{person}{Wei Wu},
  \bibinfo{person}{Xinhan Di}, {and} \bibinfo{person}{Baoquan Chen}.}
  \bibinfo{year}{2018}\natexlab{}.
\newblock \showarticletitle{{PointCNN}: Convolution on X-transformed points}.
  In \bibinfo{booktitle}{\emph{NeurIPS}}.
\newblock


\bibitem[Lipman et~al\mbox{.}(2010)]%
        {Lipman2010}
\bibfield{author}{\bibinfo{person}{Yaron Lipman}, \bibinfo{person}{Raif~M
  Rustamov}, {and} \bibinfo{person}{Thomas~A Funkhouser}.}
  \bibinfo{year}{2010}\natexlab{}.
\newblock \showarticletitle{Biharmonic distance}.
\newblock \bibinfo{journal}{\emph{ACM Trans. Graph.}} \bibinfo{volume}{29},
  \bibinfo{number}{3} (\bibinfo{year}{2010}).
\newblock


\bibitem[Liu et~al\mbox{.}(2020)]%
        {Liu2020c}
\bibfield{author}{\bibinfo{person}{Hsueh-Ti~Derek Liu},
  \bibinfo{person}{Vladimir~G. Kim}, \bibinfo{person}{Siddhartha Chaudhuri},
  \bibinfo{person}{Noam Aigerman}, {and} \bibinfo{person}{Alec Jacobson}.}
  \bibinfo{year}{2020}\natexlab{}.
\newblock \showarticletitle{Neural Subdivision}.
\newblock \bibinfo{journal}{\emph{ACM Trans. Graph. (SIGGRAPH)}}
  \bibinfo{volume}{39}, \bibinfo{number}{4} (\bibinfo{year}{2020}).
\newblock


\bibitem[Loop(1987)]%
        {Loop1987}
\bibfield{author}{\bibinfo{person}{Charles Loop}.}
  \bibinfo{year}{1987}\natexlab{}.
\newblock \showarticletitle{Smooth subdivision surfaces based on triangles}.
\newblock \bibinfo{journal}{\emph{Thesis, University of Utah}}
  (\bibinfo{year}{1987}).
\newblock


\bibitem[Loshchilov and Hutter(2019)]%
        {Loshchilov2017}
\bibfield{author}{\bibinfo{person}{Ilya Loshchilov} {and}
  \bibinfo{person}{Frank Hutter}.} \bibinfo{year}{2019}\natexlab{}.
\newblock \showarticletitle{Decoupled weight decay regularization}. In
  \bibinfo{booktitle}{\emph{ICLR}}.
\newblock


\bibitem[Maehara(2013)]%
        {maehara2013euclidean}
\bibfield{author}{\bibinfo{person}{Hiroshi Maehara}.}
  \bibinfo{year}{2013}\natexlab{}.
\newblock \showarticletitle{Euclidean embeddings of finite metric spaces}.
\newblock \bibinfo{journal}{\emph{Discrete Mathematics}} \bibinfo{volume}{313},
  \bibinfo{number}{23} (\bibinfo{year}{2013}).
\newblock


\bibitem[Martinek et~al\mbox{.}(2012)]%
        {Martinek2012}
\bibfield{author}{\bibinfo{person}{Michael Martinek}, \bibinfo{person}{Matthias
  Ferstl}, {and} \bibinfo{person}{Roberto Grosso}.}
  \bibinfo{year}{2012}\natexlab{}.
\newblock \showarticletitle{{3D} Shape Matching based on Geodesic Distance
  Distributions}. In \bibinfo{booktitle}{\emph{Vision, Modeling and
  Visualization}}.
\newblock


\bibitem[M{\'e}moli and Sapiro(2001)]%
        {Memoli2001}
\bibfield{author}{\bibinfo{person}{Facundo M{\'e}moli} {and}
  \bibinfo{person}{Guillermo Sapiro}.} \bibinfo{year}{2001}\natexlab{}.
\newblock \showarticletitle{Fast computation of weighted distance functions and
  geodesics on implicit hyper-surfaces}.
\newblock \bibinfo{journal}{\emph{Journal of computational Physics}}
  \bibinfo{volume}{173}, \bibinfo{number}{2} (\bibinfo{year}{2001}).
\newblock


\bibitem[M{\'e}moli and Sapiro(2005)]%
        {Memoli2005}
\bibfield{author}{\bibinfo{person}{Facundo M{\'e}moli} {and}
  \bibinfo{person}{Guillermo Sapiro}.} \bibinfo{year}{2005}\natexlab{}.
\newblock \showarticletitle{Distance Functions and Geodesics on Submanifolds of
  {$R^d$} and Point Clouds}.
\newblock \bibinfo{journal}{\emph{SIAM J. Appl. Math.}} \bibinfo{volume}{65},
  \bibinfo{number}{4} (\bibinfo{year}{2005}).
\newblock


\bibitem[Mitchell et~al\mbox{.}(1987)]%
        {Mitchell1987}
\bibfield{author}{\bibinfo{person}{Joseph S.~B. Mitchell},
  \bibinfo{person}{David~M. Mount}, {and} \bibinfo{person}{Christos~H.
  Papadimitriou}.} \bibinfo{year}{1987}\natexlab{}.
\newblock \showarticletitle{The discrete geodesic problem}.
\newblock \bibinfo{journal}{\emph{SIAM J. Comput.}} \bibinfo{volume}{16},
  \bibinfo{number}{4} (\bibinfo{year}{1987}).
\newblock


\bibitem[Osada et~al\mbox{.}(2002)]%
        {Osada2002}
\bibfield{author}{\bibinfo{person}{Robert Osada}, \bibinfo{person}{Thomas
  Funkhouser}, \bibinfo{person}{Bernard Chazelle}, {and} \bibinfo{person}{David
  Dobkin}.} \bibinfo{year}{2002}\natexlab{}.
\newblock \showarticletitle{Shape distributions}.
\newblock \bibinfo{journal}{\emph{ACM Trans. Graph.}} \bibinfo{volume}{21},
  \bibinfo{number}{4} (\bibinfo{year}{2002}).
\newblock


\bibitem[Panozzo et~al\mbox{.}(2013)]%
        {Panozzo2013}
\bibfield{author}{\bibinfo{person}{Daniele Panozzo}, \bibinfo{person}{Ilya
  Baran}, \bibinfo{person}{Olga Diamanti}, {and} \bibinfo{person}{Olga
  Sorkine-Hornung}.} \bibinfo{year}{2013}\natexlab{}.
\newblock \showarticletitle{Weighted averages on surfaces}.
\newblock \bibinfo{journal}{\emph{ACM Trans. Graph. (SIGGRAPH)}}
  \bibinfo{volume}{32}, \bibinfo{number}{4} (\bibinfo{year}{2013}).
\newblock


\bibitem[Paszke et~al\mbox{.}(2019)]%
        {Paszke2019}
\bibfield{author}{\bibinfo{person}{Adam Paszke}, \bibinfo{person}{Sam Gross},
  \bibinfo{person}{Francisco Massa}, \bibinfo{person}{Adam Lerer},
  \bibinfo{person}{James Bradbury}, \bibinfo{person}{Gregory Chanan},
  \bibinfo{person}{Trevor Killeen}, \bibinfo{person}{Zeming Lin},
  \bibinfo{person}{Natalia Gimelshein}, \bibinfo{person}{Luca Antiga},
  \bibinfo{person}{Alban Desmaison}, \bibinfo{person}{Andreas Kopf},
  \bibinfo{person}{Edward Yang}, \bibinfo{person}{Zachary DeVito},
  \bibinfo{person}{Martin Raison}, \bibinfo{person}{Alykhan Tejani},
  \bibinfo{person}{Sasank Chilamkurthy}, \bibinfo{person}{Benoit Steiner},
  \bibinfo{person}{Lu Fang}, \bibinfo{person}{Junjie Bai}, {and}
  \bibinfo{person}{Soumith Chintala}.} \bibinfo{year}{2019}\natexlab{}.
\newblock \showarticletitle{{PyTorch}: An Imperative Style, High-Performance
  Deep Learning Library}. In \bibinfo{booktitle}{\emph{NeurIPS}}.
\newblock


\bibitem[Pfaff et~al\mbox{.}(2020)]%
        {Pfaff2020}
\bibfield{author}{\bibinfo{person}{Tobias Pfaff}, \bibinfo{person}{Meire
  Fortunato}, \bibinfo{person}{Alvaro Sanchez-Gonzalez}, {and}
  \bibinfo{person}{Peter Battaglia}.} \bibinfo{year}{2020}\natexlab{}.
\newblock \showarticletitle{Learning Mesh-Based Simulation with Graph
  Networks}. In \bibinfo{booktitle}{\emph{ICLR}}.
\newblock


\bibitem[Pressley and Pressley(2010)]%
        {pressley2010gauss}
\bibfield{author}{\bibinfo{person}{Andrew Pressley} {and}
  \bibinfo{person}{Andrew Pressley}.} \bibinfo{year}{2010}\natexlab{}.
\newblock \showarticletitle{Gauss’ theorema egregium}.
\newblock \bibinfo{journal}{\emph{Elementary differential geometry}}
  (\bibinfo{year}{2010}).
\newblock


\bibitem[Qi et~al\mbox{.}(2017)]%
        {Qi2017}
\bibfield{author}{\bibinfo{person}{Charles~R. Qi}, \bibinfo{person}{Li Yi},
  \bibinfo{person}{Hao Su}, {and} \bibinfo{person}{Leonidas~J. Guibas}.}
  \bibinfo{year}{2017}\natexlab{}.
\newblock \showarticletitle{{PointNet++}: Deep hierarchical feature learning on
  point sets in a metric space}. In \bibinfo{booktitle}{\emph{NeurIPS}}.
\newblock


\bibitem[Qin et~al\mbox{.}(2016)]%
        {Qin2016}
\bibfield{author}{\bibinfo{person}{Yipeng Qin}, \bibinfo{person}{Xiaoguang
  Han}, \bibinfo{person}{Hongchuan Yu}, \bibinfo{person}{Yizhou Yu}, {and}
  \bibinfo{person}{Jianjun Zhang}.} \bibinfo{year}{2016}\natexlab{}.
\newblock \showarticletitle{Fast and exact discrete geodesic computation based
  on triangle-oriented wavefront propagation}.
\newblock \bibinfo{journal}{\emph{ACM Trans. Graph. (SIGGRAPH)}}
  \bibinfo{volume}{35}, \bibinfo{number}{4} (\bibinfo{year}{2016}).
\newblock


\bibitem[Raviv et~al\mbox{.}(2010)]%
        {Raviv2010}
\bibfield{author}{\bibinfo{person}{Dan Raviv}, \bibinfo{person}{Alexander~M
  Bronstein}, \bibinfo{person}{Michael~M Bronstein}, {and} \bibinfo{person}{Ron
  Kimmel}.} \bibinfo{year}{2010}\natexlab{}.
\newblock \showarticletitle{Full and partial symmetries of non-rigid shapes}.
\newblock \bibinfo{journal}{\emph{International Journal of Computer Vision}}
  \bibinfo{volume}{89}, \bibinfo{number}{1} (\bibinfo{year}{2010}).
\newblock


\bibitem[Ronneberger et~al\mbox{.}(2015)]%
        {Ronneberger2015}
\bibfield{author}{\bibinfo{person}{Olaf Ronneberger}, \bibinfo{person}{Philipp
  Fischer}, {and} \bibinfo{person}{Thomas Brox}.}
  \bibinfo{year}{2015}\natexlab{}.
\newblock \showarticletitle{{U-Net}: Convolutional networks for biomedical
  image segmentation}. In \bibinfo{booktitle}{\emph{International Conference on
  Medical image computing and computer-assisted intervention}}.
\newblock


\bibitem[Rustamov et~al\mbox{.}(2009)]%
        {Rustamov2009}
\bibfield{author}{\bibinfo{person}{Raif~M Rustamov}, \bibinfo{person}{Yaron
  Lipman}, {and} \bibinfo{person}{Thomas Funkhouser}.}
  \bibinfo{year}{2009}\natexlab{}.
\newblock \showarticletitle{Interior distance using barycentric coordinates}.
\newblock \bibinfo{journal}{\emph{Comput. Graph. Forum (SGP)}}
  \bibinfo{volume}{28}, \bibinfo{number}{5} (\bibinfo{year}{2009}).
\newblock


\bibitem[Schmidt et~al\mbox{.}(2006)]%
        {Schmidt2006}
\bibfield{author}{\bibinfo{person}{Ryan Schmidt}, \bibinfo{person}{Cindy
  Grimm}, {and} \bibinfo{person}{Brian Wyvill}.}
  \bibinfo{year}{2006}\natexlab{}.
\newblock \showarticletitle{Interactive Decal Compositing with Discrete
  Exponential Maps}. In \bibinfo{booktitle}{\emph{SIGGRAPH}}.
\newblock


\bibitem[Sethian(1999)]%
        {Sethian1999}
\bibfield{author}{\bibinfo{person}{James~A. Sethian}.}
  \bibinfo{year}{1999}\natexlab{}.
\newblock \showarticletitle{Fast marching methods}.
\newblock \bibinfo{journal}{\emph{SIAM review}} \bibinfo{volume}{41},
  \bibinfo{number}{2} (\bibinfo{year}{1999}).
\newblock


\bibitem[Shamai et~al\mbox{.}(2018)]%
        {shamai2018efficient}
\bibfield{author}{\bibinfo{person}{Gil Shamai}, \bibinfo{person}{Michael
  Zibulevsky}, {and} \bibinfo{person}{Ron Kimmel}.}
  \bibinfo{year}{2018}\natexlab{}.
\newblock \showarticletitle{Efficient inter-geodesic distance computation and
  fast classical scaling}.
\newblock \bibinfo{journal}{\emph{TPAMI}} \bibinfo{volume}{42},
  \bibinfo{number}{1} (\bibinfo{year}{2018}).
\newblock


\bibitem[Sharp and Crane(2020)]%
        {Sharp2020}
\bibfield{author}{\bibinfo{person}{Nicholas Sharp} {and}
  \bibinfo{person}{Keenan Crane}.} \bibinfo{year}{2020}\natexlab{}.
\newblock \showarticletitle{You can find geodesic paths in triangle meshes by
  just flipping edges}.
\newblock \bibinfo{journal}{\emph{ACM Trans. Graph. (SIGGRAPH ASIA)}}
  \bibinfo{volume}{39}, \bibinfo{number}{6} (\bibinfo{year}{2020}).
\newblock


\bibitem[Simonovsky and Komodakis(2017)]%
        {Simonovsky2017}
\bibfield{author}{\bibinfo{person}{Martin Simonovsky} {and}
  \bibinfo{person}{Nikos Komodakis}.} \bibinfo{year}{2017}\natexlab{}.
\newblock \showarticletitle{Dynamic edge-conditioned filters in convolutional
  neural networks on graphs}. In \bibinfo{booktitle}{\emph{CVPR}}.
\newblock


\bibitem[Solomon et~al\mbox{.}(2014)]%
        {Solomon2014a}
\bibfield{author}{\bibinfo{person}{Justin Solomon}, \bibinfo{person}{Raif
  Rustamov}, \bibinfo{person}{Leonidas Guibas}, {and} \bibinfo{person}{Adrian
  Butscher}.} \bibinfo{year}{2014}\natexlab{}.
\newblock \showarticletitle{Earth mover's distances on discrete surfaces}.
\newblock \bibinfo{journal}{\emph{ACM Trans. Graph. (SIGGRAPH)}}
  \bibinfo{volume}{33}, \bibinfo{number}{4} (\bibinfo{year}{2014}).
\newblock


\bibitem[Surazhsky et~al\mbox{.}(2005)]%
        {Surazhsky2005}
\bibfield{author}{\bibinfo{person}{Vitaly Surazhsky}, \bibinfo{person}{Tatiana
  Surazhsky}, \bibinfo{person}{Danil Kirsanov}, \bibinfo{person}{Steven~J
  Gortler}, {and} \bibinfo{person}{Hugues Hoppe}.}
  \bibinfo{year}{2005}\natexlab{}.
\newblock \showarticletitle{Fast exact and approximate geodesics on meshes}.
\newblock \bibinfo{journal}{\emph{ACM Trans. Graph.}} \bibinfo{volume}{24},
  \bibinfo{number}{3} (\bibinfo{year}{2005}).
\newblock


\bibitem[Tao et~al\mbox{.}(2019)]%
        {Tao2019}
\bibfield{author}{\bibinfo{person}{Jiong Tao}, \bibinfo{person}{Juyong Zhang},
  \bibinfo{person}{Bailin Deng}, \bibinfo{person}{Zheng Fang},
  \bibinfo{person}{Yue Peng}, {and} \bibinfo{person}{Ying He}.}
  \bibinfo{year}{2019}\natexlab{}.
\newblock \showarticletitle{Parallel and scalable heat methods for geodesic
  distance computation}.
\newblock \bibinfo{journal}{\emph{IEEE Trans. Pattern Anal. Mach. Intell.}}
  \bibinfo{volume}{43}, \bibinfo{number}{2} (\bibinfo{year}{2019}).
\newblock


\bibitem[Thomas et~al\mbox{.}(2019)]%
        {Thomas2019}
\bibfield{author}{\bibinfo{person}{Hugues Thomas}, \bibinfo{person}{Charles~R.
  Qi}, \bibinfo{person}{Jean-Emmanuel Deschaud}, \bibinfo{person}{Beatriz
  Marcotegui}, \bibinfo{person}{Fran\c{c}ois Goulette}, {and}
  \bibinfo{person}{Leonidas~J. Guibas}.} \bibinfo{year}{2019}\natexlab{}.
\newblock \showarticletitle{{KPConv}: Flexible and deformable convolution for
  point clouds}. In \bibinfo{booktitle}{\emph{ICCV}}.
\newblock


\bibitem[Tsitsiklis(1995)]%
        {Tsitsiklis1995}
\bibfield{author}{\bibinfo{person}{John~N. Tsitsiklis}.}
  \bibinfo{year}{1995}\natexlab{}.
\newblock \showarticletitle{Efficient algorithms for globally optimal
  trajectories}.
\newblock \bibinfo{journal}{\emph{IEEE transactions on Automatic Control}}
  \bibinfo{volume}{40}, \bibinfo{number}{9} (\bibinfo{year}{1995}).
\newblock


\bibitem[Velickovic et~al\mbox{.}(2017)]%
        {Velickovic2017}
\bibfield{author}{\bibinfo{person}{Petar Velickovic}, \bibinfo{person}{Guillem
  Cucurull}, \bibinfo{person}{Arantxa Casanova}, \bibinfo{person}{Adriana
  Romero}, \bibinfo{person}{Pietro Lio}, {and} \bibinfo{person}{Yoshua
  Bengio}.} \bibinfo{year}{2017}\natexlab{}.
\newblock \showarticletitle{Graph attention networks}.
\newblock \bibinfo{journal}{\emph{stat}} \bibinfo{volume}{1050},
  \bibinfo{number}{20} (\bibinfo{year}{2017}).
\newblock


\bibitem[Wang et~al\mbox{.}(2018)]%
        {Wang2018}
\bibfield{author}{\bibinfo{person}{Nanyang Wang}, \bibinfo{person}{Yinda
  Zhang}, \bibinfo{person}{Zhuwen Li}, \bibinfo{person}{Yanwei Fu},
  \bibinfo{person}{Wei Liu}, {and} \bibinfo{person}{Yu-Gang Jiang}.}
  \bibinfo{year}{2018}\natexlab{}.
\newblock \showarticletitle{{Pixel2Mesh}: Generating {3D} mesh models from
  single {RGB} images}. In \bibinfo{booktitle}{\emph{CVPR}}.
\newblock


\bibitem[Wang et~al\mbox{.}(2022)]%
        {Wang2022}
\bibfield{author}{\bibinfo{person}{Peng-Shuai Wang}, \bibinfo{person}{Yang
  Liu}, {and} \bibinfo{person}{Xin Tong}.} \bibinfo{year}{2022}\natexlab{}.
\newblock \showarticletitle{Dual Octree Graph Networks for Learning Adaptive
  Volumetric Shape Representations}.
\newblock \bibinfo{journal}{\emph{ACM Trans. Graph. (SIGGRAPH)}}
  \bibinfo{volume}{41}, \bibinfo{number}{4} (\bibinfo{year}{2022}).
\newblock


\bibitem[Wang et~al\mbox{.}(2019)]%
        {Wang2019c}
\bibfield{author}{\bibinfo{person}{Yue Wang}, \bibinfo{person}{Yongbin Sun},
  \bibinfo{person}{Ziwei Liu}, \bibinfo{person}{Sanjay~E Sarma},
  \bibinfo{person}{Michael~M Bronstein}, {and} \bibinfo{person}{Justin~M
  Solomon}.} \bibinfo{year}{2019}\natexlab{}.
\newblock \showarticletitle{Dynamic graph {CNN} for learning on point clouds}.
\newblock \bibinfo{journal}{\emph{ACM Trans. Graph.}} \bibinfo{volume}{38},
  \bibinfo{number}{5} (\bibinfo{year}{2019}).
\newblock


\bibitem[Weber et~al\mbox{.}(2008)]%
        {weber2008parallel}
\bibfield{author}{\bibinfo{person}{Ofir Weber}, \bibinfo{person}{Yohai~S
  Devir}, \bibinfo{person}{Alexander~M Bronstein}, \bibinfo{person}{Michael~M
  Bronstein}, {and} \bibinfo{person}{Ron Kimmel}.}
  \bibinfo{year}{2008}\natexlab{}.
\newblock \showarticletitle{Parallel algorithms for approximation of distance
  maps on parametric surfaces}.
\newblock \bibinfo{journal}{\emph{ACM Transactions on Graphics (TOG)}}
  (\bibinfo{year}{2008}).
\newblock


\bibitem[Wu and He(2018)]%
        {Wu2018d}
\bibfield{author}{\bibinfo{person}{Yuxin Wu} {and} \bibinfo{person}{Kaiming
  He}.} \bibinfo{year}{2018}\natexlab{}.
\newblock \showarticletitle{Group normalization}. In
  \bibinfo{booktitle}{\emph{ECCV}}.
\newblock


\bibitem[Wu et~al\mbox{.}(2020)]%
        {Wu2020}
\bibfield{author}{\bibinfo{person}{Zonghan Wu}, \bibinfo{person}{Shirui Pan},
  \bibinfo{person}{Fengwen Chen}, \bibinfo{person}{Guodong Long},
  \bibinfo{person}{Chengqi Zhang}, {and} \bibinfo{person}{S~Yu Philip}.}
  \bibinfo{year}{2020}\natexlab{}.
\newblock \showarticletitle{A comprehensive survey on graph neural networks}.
\newblock \bibinfo{journal}{\emph{IEEE Transactions on Neural Networks and
  Learning Systems}} \bibinfo{volume}{32}, \bibinfo{number}{1}
  (\bibinfo{year}{2020}).
\newblock


\bibitem[Xia et~al\mbox{.}(2021)]%
        {Xia2021}
\bibfield{author}{\bibinfo{person}{Qianwei Xia}, \bibinfo{person}{Juyong
  Zhang}, \bibinfo{person}{Zheng Fang}, \bibinfo{person}{Jin Li},
  \bibinfo{person}{Mingyue Zhang}, \bibinfo{person}{Bailin Deng}, {and}
  \bibinfo{person}{Ying He}.} \bibinfo{year}{2021}\natexlab{}.
\newblock \showarticletitle{{GeodesicEmbedding (GE)}: a high-dimensional
  embedding approach for fast geodesic distance queries}.
\newblock \bibinfo{journal}{\emph{IEEE. T. Vis. Comput. Gr.}}
  (\bibinfo{year}{2021}).
\newblock


\bibitem[Xin and Wang(2009)]%
        {Xin2009}
\bibfield{author}{\bibinfo{person}{Shi-Qing Xin} {and} \bibinfo{person}{Guo-Jin
  Wang}.} \bibinfo{year}{2009}\natexlab{}.
\newblock \showarticletitle{Improving {Chen and Han}'s algorithm on the
  discrete geodesic problem}.
\newblock \bibinfo{journal}{\emph{ACM Trans. Graph. (SIGGRAPH)}}
  \bibinfo{volume}{28}, \bibinfo{number}{4} (\bibinfo{year}{2009}).
\newblock


\bibitem[Xin et~al\mbox{.}(2012)]%
        {Xin2012}
\bibfield{author}{\bibinfo{person}{Shi-Qing Xin}, \bibinfo{person}{Xiang Ying},
  {and} \bibinfo{person}{Ying He}.} \bibinfo{year}{2012}\natexlab{}.
\newblock \showarticletitle{Constant-time all-pairs geodesic distance query on
  triangle meshes}. In \bibinfo{booktitle}{\emph{I3D}}.
\newblock


\bibitem[Xu et~al\mbox{.}(2015)]%
        {Xu2015b}
\bibfield{author}{\bibinfo{person}{Chunxu Xu}, \bibinfo{person}{Tuanfeng~Y
  Wang}, \bibinfo{person}{Yong-Jin Liu}, \bibinfo{person}{Ligang Liu}, {and}
  \bibinfo{person}{Ying He}.} \bibinfo{year}{2015}\natexlab{}.
\newblock \showarticletitle{Fast wavefront propagation {(FWP)} for computing
  exact geodesic distances on meshes}.
\newblock \bibinfo{journal}{\emph{IEEE. T. Vis. Comput. Gr.}}
  \bibinfo{volume}{21}, \bibinfo{number}{7} (\bibinfo{year}{2015}).
\newblock


\bibitem[Xu et~al\mbox{.}(2018b)]%
        {Xu2018a}
\bibfield{author}{\bibinfo{person}{Keyulu Xu}, \bibinfo{person}{Weihua Hu},
  \bibinfo{person}{Jure Leskovec}, {and} \bibinfo{person}{Stefanie Jegelka}.}
  \bibinfo{year}{2018}\natexlab{b}.
\newblock \showarticletitle{How powerful are graph neural networks?}
\newblock \bibinfo{journal}{\emph{arXiv preprint arXiv:1810.00826}}
  (\bibinfo{year}{2018}).
\newblock


\bibitem[Xu et~al\mbox{.}(2009)]%
        {Xu2009}
\bibfield{author}{\bibinfo{person}{Kai Xu}, \bibinfo{person}{Hao Zhang},
  \bibinfo{person}{Andrea Tagliasacchi}, \bibinfo{person}{Ligang Liu},
  \bibinfo{person}{Guo Li}, \bibinfo{person}{Min Meng}, {and}
  \bibinfo{person}{Yueshan Xiong}.} \bibinfo{year}{2009}\natexlab{}.
\newblock \showarticletitle{Partial intrinsic reflectional symmetry of {3D}
  shapes}.
\newblock In \bibinfo{booktitle}{\emph{ACM Trans. Graph. (SIGGRAPH ASIA)}}.
\newblock


\bibitem[Xu et~al\mbox{.}(2018a)]%
        {Xu2018}
\bibfield{author}{\bibinfo{person}{Yifan Xu}, \bibinfo{person}{Tianqi Fan},
  \bibinfo{person}{Mingye Xu}, \bibinfo{person}{Long Zeng}, {and}
  \bibinfo{person}{Yu Qiao}.} \bibinfo{year}{2018}\natexlab{a}.
\newblock \showarticletitle{{SpiderCNN}: Deep Learning on Point Sets with
  Parameterized Convolutional Filters}. In \bibinfo{booktitle}{\emph{ECCV}}.
\newblock


\bibitem[Yen et~al\mbox{.}(2007)]%
        {Yen2007}
\bibfield{author}{\bibinfo{person}{Luh Yen}, \bibinfo{person}{Francois Fouss},
  \bibinfo{person}{Christine Decaestecker}, \bibinfo{person}{Pascal Francq},
  {and} \bibinfo{person}{Marco Saerens}.} \bibinfo{year}{2007}\natexlab{}.
\newblock \showarticletitle{Graph nodes clustering based on the commute-time
  kernel}. In \bibinfo{booktitle}{\emph{Advances in Knowledge Discovery and
  Data Mining}}.
\newblock


\bibitem[Yi et~al\mbox{.}(2017)]%
        {Yi2017b}
\bibfield{author}{\bibinfo{person}{Li Yi}, \bibinfo{person}{Hao Su},
  \bibinfo{person}{Xingwen Guo}, {and} \bibinfo{person}{Leonidas~J. Guibas}.}
  \bibinfo{year}{2017}\natexlab{}.
\newblock \showarticletitle{{SyncSpecCNN}: Synchronized spectral {CNN} for {3D}
  shape segmentation}. In \bibinfo{booktitle}{\emph{CVPR}}.
\newblock


\bibitem[Ying et~al\mbox{.}(2013)]%
        {Ying2013}
\bibfield{author}{\bibinfo{person}{Xiang Ying}, \bibinfo{person}{Xiaoning
  Wang}, {and} \bibinfo{person}{Ying He}.} \bibinfo{year}{2013}\natexlab{}.
\newblock \showarticletitle{Saddle vertex graph ({SVG}) a novel solution to the
  discrete geodesic problem}.
\newblock \bibinfo{journal}{\emph{ACM Trans. Graph. (SIGGRAPH ASIA)}}
  \bibinfo{volume}{32}, \bibinfo{number}{6} (\bibinfo{year}{2013}).
\newblock


\bibitem[Ying et~al\mbox{.}(2014)]%
        {Ying2014}
\bibfield{author}{\bibinfo{person}{Xiang Ying}, \bibinfo{person}{Shi-Qing Xin},
  {and} \bibinfo{person}{Ying He}.} \bibinfo{year}{2014}\natexlab{}.
\newblock \showarticletitle{{Parallel Chen-Han (PCH)} algorithm for discrete
  geodesics}.
\newblock \bibinfo{journal}{\emph{ACM Trans. Graph.}} \bibinfo{volume}{33},
  \bibinfo{number}{1} (\bibinfo{year}{2014}).
\newblock


\bibitem[Zhang et~al\mbox{.}(2023)]%
        {zhang2023neurogf}
\bibfield{author}{\bibinfo{person}{Qijian Zhang}, \bibinfo{person}{Junhui Hou},
  \bibinfo{person}{Yohanes~Yudhi Adikusuma}, \bibinfo{person}{Wenping Wang},
  {and} \bibinfo{person}{Ying He}.} \bibinfo{year}{2023}\natexlab{}.
\newblock \showarticletitle{NeuroGF: A Neural Representation for Fast Geodesic
  Distance and Path Queries}.
\newblock \bibinfo{journal}{\emph{arXiv preprint arXiv:2306.00658}}
  (\bibinfo{year}{2023}).
\newblock


\bibitem[Zigelman et~al\mbox{.}(2002)]%
        {Zigelman2002}
\bibfield{author}{\bibinfo{person}{Gil Zigelman}, \bibinfo{person}{Ron Kimmel},
  {and} \bibinfo{person}{Nahum Kiryati}.} \bibinfo{year}{2002}\natexlab{}.
\newblock \showarticletitle{Texture mapping using surface flattening via
  multidimensional scaling}.
\newblock \bibinfo{journal}{\emph{IEEE. T. Vis. Comput. Gr.}}
  \bibinfo{volume}{8}, \bibinfo{number}{2} (\bibinfo{year}{2002}).
\newblock


\end{thebibliography}
